\documentclass[lettersize,journal]{IEEEtran}
\usepackage{amsmath,amsfonts,amssymb}
\usepackage{algorithmic}
\usepackage{array}
\usepackage[caption=false,font=normalsize,labelfont=sf,textfont=sf]{subfig}
\usepackage{textcomp}
\usepackage{stfloats}
\usepackage{url}
\usepackage{verbatim}
\usepackage{graphicx}
\usepackage{color}
\usepackage{xcolor}
\usepackage{comment}
\usepackage{svg}
\usepackage{tikz}
\usepackage{booktabs}
\usepackage{wrapfig}
\usepackage[normalem]{ulem}
\usepackage{xspace}
\usepackage{listings}
\usepackage[numbers,sort&compress]{natbib}
\usepackage[hidelinks]{hyperref}
\usepackage[capitalize]{cleveref}
\usepackage[inline]{enumitem}
\usepackage{multirow}
\usepackage{pifont}
\usepackage{balance}
\def\BibTeX{{\rm B\kern-.05em{\sc i\kern-.025em b}\kern-.08em
    T\kern-.1667em\lower.7ex\hbox{E}\kern-.125emX}}

\usepackage{multirow}
\newcommand{\ourTitle}[0]{ROAD-Waymo: A Large-Scale Action Awareness Dataset for Autonomous Driving}

\newcommand{\supmat}{\mbox{\textbf{Sup.~Mat.}}\xspace}

\newcommand{\etc}{\textit{etc}\xspace}
\newcommand{\etal}{\textit{et al.}\xspace}
\newcommand{\ie}{\textit{i.e}.\xspace}

\newcommand{\eg}{\textit{e.g.,}\xspace}

\newcommand{\roadwaymo}[0]{ROAD-Waymo\xspace}
\newcommand{\waymo}[0]{Waymo\xspace}
\newcommand{\road}[0]{ROAD\xspace}
\newcommand{\roadpp}[0]{ROAD++\xspace}
\newcommand{\roadr}[0]{ROAD-R\xspace}
\newcommand{\uda}[0]{UDA\xspace}
\newcommand{\GUR}[1]{\textcolor{RoyalBlue}{Gurkirt:#1}}

\usepackage{pifont}
\usepackage{xcolor}
\newcommand{\cmark}{\color{green}\ding{51}}
\newcommand{\xmark}{\color{red}\ding{55}}

\newcommand{\sk}[1]{\textcolor{brown}{ Salman: #1}}

\newcommand{\remove}[1]{}%{\textcolor{gray}{\sout{#1}}}

\author{%
\textbf{Salman Khan}$^1$ \quad \textbf{Izzeddin Teeti}$^1$ \quad \textbf{Reza Javanmard Alitappeh}$^2$ \quad \textbf{Mihaela C. Stoian}$^3$\\
\textbf{Eleonora Giunchiglia}$^4$ \quad \textbf{Gurkirt~Singh}$^5$ \quad \textbf{Andrew~Bradley}$^1$ \quad \textbf{ Fabio~Cuzzolin}$^1$\\
$^1$Oxford Brookes University \quad $^2$MAZUST \quad $^3$ University of Oxford \quad $^4$Imperial College London \quad $^5$ETH Zurich \\
}

\begin{document}

\title{\ourTitle}

\maketitle

\begin{abstract}

% Autonomous vehicles (AVs) offer the potential for enhancements in road safety and economic benefits. Recent developments in AV perception systems, coupled with the widespread availability of datasets for training, enable the vehicle to 'see' with a high degree of accuracy. 
%% Movitvatio for creating large dataset
% Autonomous vehicles (AVs) offer the potential to improve road safety. 
Autonomous Vehicle (AV) perception systems require more than simply seeing, via e.g., object detection or scene segmentation.
% they needs a more 
%human-like 
They need a holistic understanding of what is happening within the scene for safe interaction with other road users.
%Seldom 
Few datasets exist for the purpose of developing and training algorithms to
%understand 
comprehend
the actions of other road users.
% , which is %fundamental 
% crucial
% to understanding road behaviour. 
\iffalse
Those that do exist are limited in size and scope, often restricted to one city,
while an
%- while the 
AV must be able to safely operate in the %complete variety 
full spectrum of 
%operating 
domains and situations
%which the vehicle 
it may encounter. 
\fi
%% Dataset is large
This paper presents \roadwaymo, an extensive dataset for the development 
and benchmarking
of techniques for 
%agent detection, action/event detection, and location detection - 
agent, action, location and event detection in road scenes,
provided as a layer upon the (US) Waymo Open dataset. Considerably larger and more challenging than any existing dataset (and encompassing multiple cities), it comes with
%provides 
198k annotated video frames, 54k agent tubes, 3.9M bounding boxes and a total of 12.4M labels. 
The 
integrity of the dataset has been confirmed and enhanced %by using 
via a novel annotation pipeline
designed for automatically identifying violations of requirements specifically designed for this dataset. 
% The scenarios provided are considerably more challenging than other datasets, providing the next level of complexity to challenge even the highest-performing algorithms. 
% It encompasses multiple cities, in a range of (labelled) weather conditions - thus facilitating the development of solutions. 
%The resulting dataset 
As ROAD-Waymo is compatible with the original (UK) \road\cite{singh2022road} dataset, it
% thus forming the basis for a new \roadpp family of datasets and challenges, 
provides the opportunity to tackle domain adaptation %tasks 
between real-world road scenarios in different countries within a novel benchmark: ROAD++. %(currently, UK$\Leftrightarrow$USA). 
%Finally, 
% This represents the first attempt to undertake these automated checks on a dataset of this nature. 
% We provide training code and baseline results for agent detection, action detection, location detection, combined agent \& action detection, and event detection, along with preliminary domain adaptation experiments. 
Dataset and code are available at: Dataset~\footnote{\url{https://github.com/salmank255/Road-waymo-dataset}}, Code~\footnote{\url{https://github.com/salmank255/ROAD_Waymo_Baseline}}.

% \AB{Discuss what results/conclusions/performance metrics to add here}
% \\
% \\
% \\
% \\
\end{abstract}

\section{Introduction}

Several large-scale, comprehensive AV training datasets have recently been released \citep{wang2025omnidrive,cordts2016cityscapes,KITTI2021,yu2020bdd100k,sakaridis2021acdc,wilson2021argoverse,RobotCarDatasetIJRR}, leveraging multiple sensors (\eg cameras, LiDAR, GPS, \etc) for the purposes of object detection, semantic segmentation, object tracking and trajectory forecasting.
% \textbf{Izzeddin \& rj pls list a few here e.g. Waymo etc... mention datset size e.g. how many frames / GB etc} \izz{I added them to the datasets section \ref{sec: av_datasets}} \AB{Okay, but we need a brief bit of info here as an overview. The detail can follow later}. 
% However, these existing datasets lack 
%labelling 
% specifically designed for the purpose of event awareness of all road users. 
However, a safe
%Safe 
interaction with other road users 
%is a challenging task - requiring 
requires much more than simply \textit{seeing} things: rather, it requires understanding actions and events being conducted or happening in the scene. 
To this end, \citep{singh2022road} introduced the ROAD dataset, which provides multi-label annotations of agents and actions for event awareness in autonomous driving. 
% Based upon the Oxford RobotCar dataset \citep{RobotCarDatasetIJRR}, 
%This dataset does, however, 
ROAD represents a starting point along the path to enable AVs to understand events. It features some degree of diversity in terms of weather conditions (including bright sunshine, wet roads, rain, snow and fog) and %while being
% \AB{@Salman pls edit / delete as appropriate}. GURKIRT: Looks good
%Though the size of the dataset is 
%relatively small,
%does provide 
can be used as
a benchmark for object/agent detection, action detection and event detection in a natural multi-label setting.
%It was adopted as the subject of the 
This was indeed the case for the ICCV 2021 ROAD workshop and challenge,\footnote{\url{https://sites.google.com/view/roadchallangeiccv2021/challenge}} which saw a combined 200+ entries for the aforementioned tasks. The momentum continued at the ECCV 2023 ROAD++ Workshop 7,\footnote{\url{https://sites.google.com/view/road-eccv2024/home}} where participation surged to 350+ submissions from 39 teams, a 75\% increase highlighting the growing adoption of event-aware frameworks in autonomous driving research. Since then, %if anything, 
the problem of situation awareness for AVs has gained even more traction, as exemplified by the works~\citep{giunchiglia2022road,yu2022argus}, with the former even being at the core the NeurIPS 2023 ROAD-R Challenge.\footnote{\url{https://sites.google.com/view/road-r/}}

% \begin{figure}[t]
%     \centering
%     \subfigure[]{\includegraphics[height = 3.5cm]{figures/sample1.jpg}}
%     \subfigure[]{\includegraphics[height = 3.5cm]{figures/sample1.png}}
% \caption{Exemplary annotations of road users in ROAD-Waymo style, explained from the (ego) viewpoint of an AV. Each road user is annotated with three distinct labels: agent, action, and location.}
% \vspace{-3mm}
% \label{fig1:theme}
% \end{figure}
\begin{figure}
    \centering
    \includegraphics[width=1\linewidth]{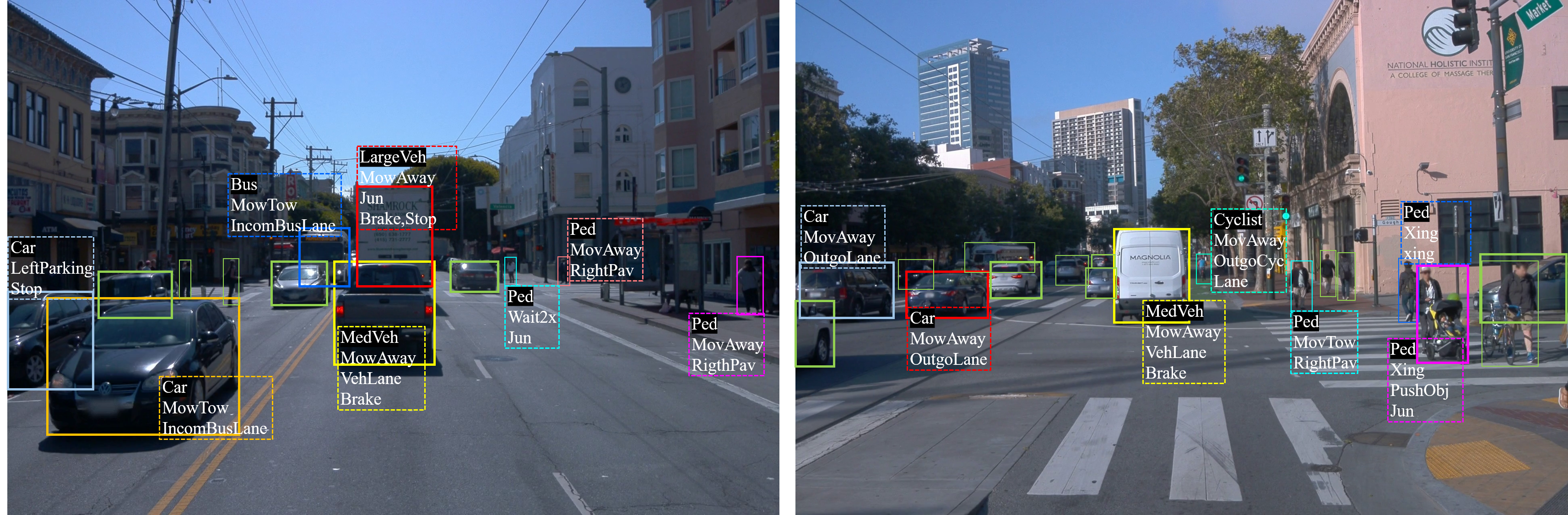}
    \vspace{-3mm}
    \caption{Exemplary annotations of road users in ROAD-Waymo style, explained from the (ego) viewpoint of an AV. Each road user is annotated with three distinct labels: agent, action, and location. Only agents with a distinct status are labeled to avoid confusion from overlapped labels.}
    \vspace{-3mm}
    \label{fig:theme}
\end{figure}

% \begin{figure}
%     \centering
%     \includegraphics[width=0.59\linewidth]{figures/theme_fig.png}
%     \vspace{-3mm}
%     \caption{Exemplary annotations of road users in ROAD-Waymo style, explained from the (ego) viewpoint of an AV. Each road user is annotated with three distinct labels: agent, action, and location.  }
%     \vspace{-3mm}
%     \label{fig:theme}
% \end{figure}

% \begin{figure}
%     \centering
%     \footnotesize
%     \includegraphics[width=0.9\linewidth]{figures/da_res.pdf}
%     \caption{Domain adaptation results 
%     (mAP, \%)}
%     \label{fig:da_res}

% \end{figure}

Despite its success, ROAD contains only 22 ($\sim\!8$-minutes long) videos, and hence remains a small-scale dataset, lacking in both scale and true diversity to allow the further demonstration and testing of the benefits of event awareness. Indeed, ut is limited to scenes from just one, relatively small city - Oxford, UK (population 160,000) - with a high density of narrow streets, relatively low-speed driving on few routes.
\iffalse
(selected to work on the % repetitively for the 
SLAM~\citep{durrant2006simultaneous} problem, rather than for situational awareness).
\fi
% , and only nearby agents labelled. 
This makes it suitable for early stage development, but is significantly less challenging than real-world driving in complex environments.

This paper introduces \roadwaymo, which tackles the diversity and scale issues of ROAD~\citep{singh2022road}, building upon the Waymo-Open dataset~\citep{sun2020scalability} by adding 
dense spatiotemporal annotations. The annotations include bounding boxes around each relevant agent in the scene together with their agent, actions and locations labels. The labelling process was based upon the \emph{road event = $\{$Agent, Actions, Locations$\}$} protocol {(illustrated in Figure \ref{fig:theme})} proposed by~\cite{singh2022road}, and the results are illustrated in Figure \ref{fig:theme}. Adding such extensive labelling to the $52K$ tracks inherited from the Waymo dataset resulted in $4.1M$ bounding boxes with $4.1M$ agent,  $4.3M$ action, and $4.3M$ location label instances. %ROAD-Waymo hence allows researchers and practitioners to study agent, action and event detection in a real world setting. 
%As such annotations are expensive, we inherited $52K$ tracks from the Waymo dataset, adding extensive labelling %in a multi-label setting, 
% based upon the \emph{road event = $\{$Agent, Actions, Locations$\}$} protocol {(illustrated in Figure \ref{fig:theme})},
% resulting in $4.1M$ bounding boxes with $4.1M$ agent,  $4.3M$ action, and $4.3M$ location label instances. 
% The larger number of action and location labels is due to the multi-label nature of action and location themselves {\color{red}this sounds a bit vague}, which had to be annotated for each box independently.  
 Additionally, \roadwaymo represents a complementary and compatible dataset to ROAD, and together they provide an extensive playground 
%to enable 
for the development of advanced perception systems for autonomous vehicles and the tackling of novel problems, with a focus on domain adaption (in particular, between cities in the UK and the US) for agent, action and event detection tasks.

\paragraph{Contributions.}
The contributions of this work are:

\begin{enumerate}
% [leftmargin=0.5cm]
% \vspace*{-0.05cm}
% \setlength\itemsep{0.1mm}
\item \textbf{\roadwaymo.} A large-scale multi-label dataset for action and event detection in autonomous vehicles. Provided as an extension to the Waymo AV dataset~\citep{sun2020scalability}, it is considerably more comprehensive and challenging than any existing such benchmark, including ROAD. %FAB: such as? be specific
This dataset presents a series of unique properties:

\begin{itemize}
    \item \textbf{Scale.} \roadwaymo contains $4.3M$ action labels, making it $7$ times larger than the original \road dataset, and the largest known dataset for this purpose. At the same time, it extends the Waymo dataset by providing a total of $12.7M$ additional labels.
    % {\color{red} Can we say this? Someone pls add the numbers above}
    
    \item \textbf{Multi-City.} The dataset encompasses 4 cities from several US states, providing a plethora of different driving scenarios in a range of operating domains.

% \item \textbf{Variable Weather:} The dataset provides a variety of labelled weather conditions, thus enabling the use of the dataset for weather-shift and augmentation purposes. {\color{red}but we are not using this}\sk{then shall we add it in the complexity rather than a separate contribution? }

\item \textbf{Complexity.} The labelling provided is highly comprehensive - actors are labelled at great distances from the ego vehicle, with many active participants in each scene. Several complex scenes contain the presence of emergency vehicles (11 times more than the \road), six weather conditions (two times of \road), many lanes, busy roads, pedestrians and so on.
% {\color{red}we should mention emergency vehicles presence}
Thus, \roadwaymo provides a considerably more challenging perception testbed 
%for a perception system 
than any other dataset of this kind.

\item \textbf{Verified Annotations.} Importantly, \roadwaymo is the first AV dataset guaranteed to be compliant with a set of commonsense constraints. Each annotation was automatically verified to be compliant with 251 domain-dependent logical requirements elicited for this task.
\end{itemize}

\item \textbf{ROAD++ domain adaptation framework.} The new \roadwaymo dataset is fully compatible with the original \road data. Together, they form the basis for a  
% the new \roadpp family of datasets and challenges for 
%the following:
%\begin{itemize} \item 
benchmarking framework for domain adaptation in autonomous driving {(between two countries UK and the US and their 5 different cities)}, termed ROAD++.
% {\color{red} be specific about what domain adaptation tasks, city-city etc}
% {\color{blue}AB: ...and weather conditions?}

\item \textbf{Comprehensive Tasks and Baselines.} Experiments were conducted to provide a benchmark for a range of varied perception tasks listed below:
\begin{itemize}
% \vspace*{-0.15cm}
% \setlength\itemsep{0.1mm}
    \item Object/Agent, Action, and Event detection tasks,
    % (Triplet detection (object-action-location))
    \item Cross-country datasets and cross-city adaptation tasks, and
    % {\color{red} should we be more specific?}
 \item Neurosymbolic prediction tasks, where we used the requirements to guide the training of the neural models to ensure compliance with commonsense rules.
% , and many are very small within the active participants in each scene
\end{itemize}
For each of the above tasks, we also provide strong baselines against which novel models can be compared.

\end{enumerate}

%\qheading{Tasks and Baselines:} 
% {\color{red}this part seems a bit rushed; also, I am not sure they should be listed as contributions? Maybe best to describe the tasks and baselines underneath the list} 

% {\color{red}I note that there is no paper outline} - \sk{I think all of the most recent conference papers avoid paper outline}

%\item \textbf{Tasks and Baselines:} Experiments were conducted to provide a benchmark for the following perception tasks; Object/Agent detection; Event detection (Triplet detection (object-action-location)); Domain adaptation (for the above tasks); Neuro-symbolic prediction enhancement: Utilising domain knowledge to guide training of the neural models.

% , and many are very small within the active participants in each scene

%%%%%%%%%%%%%%%%%%%%%%%%%%%%%%%%%%%%%%%%%%%%%%%%%%%%%%%%%%%%%%%%5

\section{Related Work}

\textbf{Autonomous Driving Datasets}. %% IZZI
%\label{sec: av_datasets}
%The field of 
Autonomous vehicle perception has recently made significant progress, thanks to modern benchmarks that offer more diverse environments and agent categories. %than older datasets. 
For instance, previous datasets like NGSIM-180 \citep{COIFMAN2017362} and highD \citep{highDdataset} captured cars on highways using drones and surveillance cameras, while PIE~\citep{Rasouli2019pie} and JAAD \citep{kotseruba2016joint} focused solely on pedestrian annotations using a single ego-vehicle camera. KITTI \citep{KITTI2021} was 
%one of the first datasets 
among the first to provide multimodal inputs, including camera frames and LiDAR point clouds, and annotations for both cars and pedestrians. Further down the line, Lyft \citep{houston2020one}, Waymo \citep{ettinger2021large}, nuScenes \citep{caesar2020nuscenes}, Argoverse \citep{wilson2021argoverse}, Argoverse 2 \citep{wilson2023argoverse}, and Zenseact \citep{Alibeigi_2023_ICCV} provide large-scale, comprehensive annotation, HD-maps, odometer information and additional labels for other agents and environmental factors like weather and lighting conditions. A more recent dataset Rank2Tell \citep{sachdeva2024rank2tell} is designed for tasks involving both ranking the importance of various elements in a driving scenario and explaining those rankings. 

While these works highlight the shift toward holistic environment modeling, they lack explicit action-event annotations like ROAD-Waymo. Indeed, their focus is on object detection, semantic segmentation or multi-object tracking tasks, lacking labels for high-level action understanding. Thus, there is no real competitor to our ROAD++ framework. A comprehensive comparison between the state-of-the-art datasets is shown in Table \ref{tab:datasets_comp}, with \roadwaymo and \roadpp added for comparison.

\begin{table*}[t]
\caption[Features of the recent and most widely used AV datasets.]{Features of the state-of-the-art autonomous vehicle datasets. ``Man." stands for manual annotation, ``Auto'' stands for automatic annotation while ``SemiAuto'' stands for an hybrid annotation pipeline. Hyphens ``-" indicate that information is not available. For every row (aside the first) we highlight in bold the highest number, which denotes the richest dataset along that particular dimension.}
\label{tab:datasets_comp}
\resizebox{\textwidth}{!}{%
% \rowcolors{2}{lightblue}{white}
\begin{tabular}{lcccccccccccc}
\toprule
\textbf{Dataset} & KITTI  & PIE & Lyft & nuScenes & Waymo-OM & LOKI & ROAD & Argoverse 2 & Zenseact &  Rank2Tell & \textbf{\roadwaymo}& \textbf{\roadpp}\\ \midrule
\textbf{Year of Release}& 2012  &  2019 & 2019  & 2019 & 2021  & 2021  & 2021  & 2023 & 2023 & 2024 & 2024 & 2024 \\
\textbf{Number of Classes}     & 3     & 1     &  3   & 10     & 3     & 8    &  \textbf{43}   & 30 & 37  & 33 & \textbf{43}   & \textbf{43} \\
\textbf{Night/Rain}     & \xmark/\xmark  & \xmark/\xmark & \cmark/\cmark & \cmark/\cmark & \cmark/\xmark & \cmark/\cmark & \cmark/\cmark 
& \xmark/\xmark  &  \xmark/\cmark &  \xmark/\xmark & \cmark/\cmark & \cmark/\cmark \\
\textbf{Number of Cities}     &   1     & 1    & 1     & 2   & 6    & -  & 3  & 6 & \textbf{14}  & 1 & 3    &  4\\
\textbf{Annotated Frames (k)}& 15   & 293    & 46   & 40  & 400     &  41   & 22 & \textbf{27k}  & 100 & 23 & 198 & 320 \\
\textbf{Location Label} & \xmark      & \xmark  & \xmark & \xmark & -  &  \cmark   &  \cmark  &  \xmark & \xmark & \cmark & \cmark & \cmark \\
\textbf{Action Label}   & \xmark   & \cmark    & \xmark  & \cmark  & \cmark    & \cmark  & \cmark   & \xmark  &  \xmark &  \xmark & \cmark & \cmark \\
%\textbf{??}    & \xmark  &   & \xmark   & \cmark & \cmark  & \cmark  & \xmark  &   &  \cmark   & &  \\
%\textbf{Multi-agent Evaluation}  & \xmark   & \cmark  & \xmark  & \cmark  & \cmark  &  \cmark &  \cmark  &   &  &  & \cmark  &  \cmark  \\
% \textbf{Sampling Rate (Hz)}& 10  &  30   & 10   & 2   & 10   & 5   & 12  &   10  &  &     &  \\
\textbf{Total Time (h)}    & 1.5     & 6 & \textbf{1157} & 5.5  & 574 &  2.3   &  3  &  4.2  & 55.6 &  0.6  & 5.5 & 8.5 \\
\textbf{Download Size (Gb)}& 354    & 22   & 48  & \textbf{1400} & -   &  4.2 &   16 & 58  & - & -  & 18 & 34 \\
\textbf{Annotation}        & Man. & Auto & Man.& Auto & Man.& Man. &  Auto & Man. & Man.  & Man. &  SemiAuto   &   SemiAuto \\  \bottomrule

\end{tabular}%
}
\vspace{-3mm}
\end{table*}

\paragraph{Action and Event Detection Datasets.} % REZA
% \textbf{Start on general datasets in all domains, then move and focus on road scene specifi data}
% In order to develop a detailed scene understanding, it is necessary to train ML models on datasets which have been specifically designed for this purpose. 
An action is a process - and, thus, temporal in nature. As such, datasets focusing on actions require detailed annotations in terms of both action labels and the localisation of the actions of interest as they evolve over time. 
% Therefore, datasets require frame-wise labelling which tracks the agents in the temporal domain - which makes the annotation extremely time-consuming. 
Thus, the number of video datasets~\citep{liang2025videvent,soomro2012ucf101,ava2017gu,daly2016weinzaepfel,J-HMDB-Jhuang-2013} for action detection is much smaller than for other video tasks, e.g., action recognition~\citep{kay2017kinetics}.
% \AB{As I understand it, action recognition datasets simply contain one label for the whole clip? Whereas activity recognition includes labels for different actors, with a location, and the action can change over time? So Kinetics and monfortmoments are only clip-level annotations? @rj pls confirm} \rj{let say, \textit{action detection} refers to the former and \textit{action recognition} to the latter one.}
%
% Several action recognition datasets have been proposed in different domains. 
% Two of the largest datasets action recognition are kinetics \citep{kay2017kinetics} with 600 classes of actions and Moments \citep{monfortmoments} which contains people, animals, objects or natural phenomena in one million 3 second video clips.
%
% In 'something-something'\citep{goyal2017something} 174 complex human actions are considered in the daily interaction with surrounding objects.
%
% ActivityNet~\citep{caba2015activitynet} and Charades~\citep{sigurdsson2018charadesego} are also two other well-known datasets in action recognition, however, they do not contain spatiotemporal properties of the actions.
%
% Another category of the data set is common with \roadpp in the sense of providing spatial and temporal annotation of human actions, i.e., J-HMDB-21 \citep{J-HMDB-Jhuang-2013}, UCF24 \citep{soomro2012ucf101}, LIRIS-HARL \citep{liris-harl-2012}, DALY \citep{daly2016weinzaepfel} and AVA \citep{ava2017gu} , however, most of them are limited in term of the presence of different source domain. 
%% GURKIRT: Condensed it; We don't need such and expensive description of recognition dataset
AVA \citep{ava2017gu} is currently the largest action detection dataset, with 1.6M labels. Still, it suffers from a relative sparsity of the annotation, at one frame per second. 
% {\color{red} doublecheck correctness}.
%Introduced in recent work, 
MultiSport is a new multi-person dataset with spatiotemporally localised sports actions \citep{li2021multisports}, 
%Compared to earlier benchmarks, it 
featuring higher diversity, denser annotation and improved quality. More recently, \cite{Biparva2022} considered the detection of car-lane-change-events as a video action recognition problem. 
% Such that the AV will predict this task to avoid any potential accidents. 
% They applied different models, i.e. I3D and slowfast architectures, with two streams: video data and optical flow motion data. 
In the same spirit, but for anomaly detection purposes, is the contribution of \cite{pmlr-v164-wiederer22a}.
% introduced a dataset that provides agent trajectories (\ie position of the cars) over the frames. Each trajectory has a label from 12 classes of defined anomalies \ie aggressive overtaking, pushing aside, right/left spreading. 
TITAN~\citep{malla2020titan} provides annotations for pedestrians' actions present in the road scene, but not for other road agents. PIE~\citep{Rasouli2019pie} also annotates 6 action classes for each road user for pedestrian behaviour anticipation. The annotation is focused on the pedestrian and the vehicles/traffic lights they interact with, and is thus once again quite sparse. All these datasets are promising and relevant: however, they are either sparsely annotated or limited in the number of labels and types (e.g., action or location) of labels, or both.
% \textcolor{red}{@Izz confirm number of action classes for PIE}
% Similarly, for anomaly detection (single class) in \citep{yao2022dota}, authors proposed an unsupervised approach for predicting future locations of cars wi   vv k,   x  b n  lk .mn nmm  km n nnvthin a short horizon to detect anomaly i.e. accidents.
To the best of our knowledge, \road \citep{singh2022road} and now ROAD-Waymo are the only dataset in which the actions of all road users are densely annotated, including agent, action, and location labels. % for each road user.
% , i.e. human and vehicle, stressing the dynamical aspect of events and the relationship between distinct but correlated events.
% \AB{Need to add the limitations of \road here}
\iffalse
In this work, we present \roadwaymo: generated by extending \road annotation to \waymo Open dataset~\citep{sun2020scalability} with complete compatibility, thus, scaling \road by over 4.3M action and 3.1M event labels.
\fi
% at least 5 times for some type of labels and more for others.
% \textcolor{red}{AB: Here we say 5 times, above it's 7, somewhere else 20 was mentioned!}

% the number of \road 
% We extend the \road dataset by adding more samples of actions with shorter duration of the agents in road environment. Therefore, in the sense of action detection, we extend the number actions instances in our new proposal.
% \rj{Salman, may you please a sentence including statistic or any advantage point in comparison to the other action detection dataset}

\paragraph{The \road Dataset.}
% \AB{This section should begin with a review of what \road actually is - discuss what is annotated, how many frames, say that it's based upon the robotcar dataset, etc. Also needs to mention the baselines that original \road provides.}
% \Mihaela{Gurkirt please check this subsection}
%{\color{red} question: do we really need this section, after already mentioning ROAD several times above?}
\iffalse
\roadpp is part of a larger-scale effort to increase the situational awareness capabilities of self-driving systems and is based on 
% the pioneering work in this direction, 
\fi
The ROad event Awareness Dataset for Autonomous Driving (\road)~\citep{singh2022road} was originally designed to test event and action recognition and detection, being the first work to shift the focus from actions performed by human bodies to actions and events of agents. 
\road contains 22 $\sim\!8$-minute long videos from the Oxford RobotCar Dataset~\citep{RobotCarDatasetIJRR} (OxRD), summing up to 122K frames. 
% OxRD was produced by the Oxford University's Robotics Institute and captures over 100 traversals of a 100 km route through Oxford, UK, within about a year, which allowed for recording different weather conditions and illuminations, but also other scene changes such as construction and seasonal effects.
% And while OxRD, at the time, contained different data types such as pointclouds and RGB images suitable for object detection tasks, they lacked support for event understanding.
% To this end, 
\road was introduced to provide event awareness  capabilities, where each agent in the scene is annotated with road events for each bounding box, specifying the type of agent present in the bounding box, its action(s) and its location(s).
The new annotations allowed for introducing new tasks such as spatiotemporal (i) agent detection, (ii) action detection, (iii) location detection, (iv) agent-action detection, (v)
road event detection, and (vi) temporal segmentation of AV actions, all of which can be tested at frame and video levels. Moreover, these annotations can be leveraged for potential tasks, \eg multi-object tracking, trajectory forecasting, action anticipation/prediction \etc. 

\paragraph{Action Detection Methods.} % REZA/SALMAN
%The field of 
Action detection encompasses two main classes of approaches. The first class \citep{faure2023holistic,fan2021multiscale,pan2021actor,tong2022videomae} involve using an auxiliary human detector to localise actors in keyframes, followed by action classification. These types of methods rely on models like Faster RCNN-R101-FPN \citep{ren2015faster} for human detection and RoIAlign \citep{he2017mask,khan2021spatiotemporal} for feature extraction. However, %this approach overlooks
they may overlook
contextual and interaction information outside the bounding box. The second class of approaches utilises end-to-end action detectors \eg \citep{alzahrani2025yolo,chen2021watch,kopuklu2019you,sun2018actor}, which employ a single model to perform action detection. These methods simplify training by jointly training actor proposal networks and action classification networks. However, they also suffer from fixed RoI feature sampling. Recent advancements include one-stage action detectors such as MOC \citep{li2020actions} and TubeR \citep{zhao2022tuber}, which have their own limitations such as relying more on appearance features and neglecting multi-scale information.
% \textcolor{red}{Add citations above} -- done

% In general, we have offline \citep{chen2019relation,liu2021multi} and online \citep{soomro2016predicting,singh2017online,behl2017incremental,kalogeiton2017action,li2020actionsas,yang2019step, Yang_2022_CVPR} action detection methods which  demonstrate acceptable performance. While in the former one all frames of a clip can be used for action detection, in the latter we do not necessarily have the next frame to use in the detection procedure.
% %
% In our application, given the incremental processing needs of an autonomous vehicle, we are dealing with online action detection using visual sensors ~\citep{singh2017online}.
% %
% Dataset UCF-101-24 is the main benchmark for online action detection research, as it provides annotation in the form of action tubes and every single frame of the untrimmed videos in it is annotated (unlike AVA~\citep{ava2017gu}, in which videos are only annotated at one frame per second).

Focusing on action detection in autonomous driving, \citep{Biparva2022,yao2022dota,khan2024hybrid} studied the trajectory and complex activities of cars in road scenarios. In \citep{Biparva2022} the focus was on detecting a specific action performed by a car - namely ``lane change". To achieve this, models such as Inflated 3D-ConvNet (I3D)~\citep{carreira2017quo}, Spatiotemporal Multiplier ConvNet and Slowfast~\citep{feichtenhofer2019slowfast} were trained to identify and classify the lane change action. %The work 
\citep{pmlr-v164-wiederer22a} tackled anomaly detection for multiple actions in driving scenarios via a spatiotemporal graph auto-encoder, developed to learn normal driving behaviors using the trajectories of vehicles as input to the model.
%Inspired by end-to-end action detectors, in this study we also adopt 3D-RetinaNet \citep{singh2022road} for road event detection. 
Beyond the more standard action detection methods, \citep{liao2025diffusiondrive} introduced DiffusionDrive, an end-to-end autonomous driving framework leveraging truncated diffusion models for trajectory planning. It achieves state-of-the-art performance on nuScenes and Waymo, emphasising the growing role of generative models in complex driving scenarios. Concurrently, \citep{yang2025driving} proposed a vision-centric Occupancy World Model for 4D occupancy forecasting and planning, addressing spatiotemporal scene understanding beyond traditional bounding boxes. 

%%%%%%%%%%%%%%%%%%%%%%%%%%%%%%%%%%%%%%%%%%%%%%%%%%%%%%%%%%%%%%%%%%%%%%
% \subsection{Domain Adaptation Datasets: Gurkirt}
% Mention general Domain Adaptation, quickly focus on ADriving

\paragraph{Unsupervised Domain Adaptation.} %%% GURKIRT
Unsupervised domain adoption models are trained on a labelled source domain and adapted to an unlabelled target domain. 
There has been considerable progress towards bridging the domain gaps in major computer vision problems, including image classification \citep{ganin2016domain,long2018conditional}, semantic segmentation \citep{hoffman2016fcns,hoyer2022daformer}
and object detection \citep{chen2021scale,li2022sigma,li2022cross}.
Semantic and panoptic segmentation~\citep{saha2023edaps} are the most studied tasks for domain adaptation in autonomous driving setting~\citep{hoyer2022daformer}. Mostly, these problems are conducted in synthetic-to-real settings, where the source domain is synthetic (\eg SYNTHIA~\citep{ros2016synthia}) but the target domain is real (\eg~\citep{cordts2016cityscapes}).
However, few works consider real-to-real settings, and mostly in subdomain settings such as Day-to-Nighttime, Clear-to-Adverse-Weather ~\citep{hoyer2022daformer,wu2021dannet,hoyer2023domain}.
% \textcolor{blue}{In}.   
% Similarly, an autonomous driving system should be able to adapt from one driving environment to another. 
%%%---------------------------
Our proposed \roadpp domain adaptation framework (making use of both \road and \roadwaymo) %{\color{red} we have not mentioned ROAD++ as a project; I note that this has implications for double blind reviewing; shall we just say "coupling ROAD and ROAD-Waymo provides the opportunity?" or we introduce the ROAD++ setting under point 7, but without claiming both datasets are ours}, 
%we have the opportunity 
enables
domain adaption and generalisation experiments between real datasets, considering both city 
%locations and weather conditions. 
and country locations.
To the best of our knowledge, \roadpp is the first real-to-real domain adaption benchmark for event awareness spanning two countries (UK \textit{vs.} USA) and four different urban settings with drastically different road environments, \eg left-hand \textit{vs.} right-hand drive, narrow \textit{vs.} wide roads \etc.

\section{\roadwaymo Dataset}

% \subsection{Introduction to what the new dataset actually 
% provides over and above other existing datasets: Gurkirt}
\roadwaymo is a new, large-scale dataset providing a considerably more detailed, comprehensive and challenging playground for the development and test of advanced perception systems for autonomous vehicles. It has been designed to ensure compatibility with the existing \road (UK) dataset, and thus carries forward the same annotation strategy (see, \eg Figure \ref{fig:theme}) - this time applied to the popular Waymo dataset. It contains scenes from multiple US cities, with 8 times the number of annotated agent tubes when compared to \road.
Importantly, for the first time a novel automated logical cross-checking process helps to ensure the integrity of the complex multi-label dataset.

% Autonomous driving has seen a lot of progress in recent year with the help of real-world large-scale datasets ~\citep{caesar2020nuscenes,wilson2021argoverse,ettinger2021large,RobotCarDatasetIJRR,sun2020scalability,yu2020bdd100k,KITTI2021,cordts2016cityscapes} and synthetic datasets~\citep{Richter_2016_ECCV,ros2016synthia,gaidon2016virtual}. Most of the above datasets deal with computer vision problems like semantic segmentation, objection detection, tracking, and synthetic to real domain adaption. However, to drive cars on city streets, problems like intention estimation, trajectory forecasting, and pedestrian behaviour modelling become very important. Datasets like TITAN~\citep{malla2020titan}, PIE~\citep{Rasouli2019pie} have made strides in a limited manner, on the other hand, \road~\citep{singh2022road} approach this problem from holistic decision-making based on the event, intent, action, location of all the road agents in the scene. 
% In this work, we extend the idea of \road by scaling it to multiple US cities, with \textcolor{black}{8 times} more agent tubes annotated compared to the original \road dataset. 
% \GUR{I am not sure if I should explore domain adaptation here. Usually synthetic to really help but real-2-real is much hard to make work.}

%%%%%%%%%%%%%%%%%%%%%%%%%%%%%%%%%%%%%%%%%%%%%%%%%%%%%%%%%%%%%%%%%%%%%%%%%%%%%%%%%%%%%%%%

\begin{table*}
\centering
\footnotesize
\caption{Comparison of \road (Oxford, UK) with the cities of \roadwaymo dataset.}
% \vspace{-3mm}
\label{tab:rpp_stats}
\scalebox{0.95}{\begin{tabular}{l|l|lllll}
\toprule
Dataset             & \road & \multicolumn{4}{c}{\roadwaymo}  \\ 
Cities              & Oxford, UK    & Phoenix, USA & San Francisco, USA & Other & Total          \\ \midrule
Total duration (h)  & 3    & 2. (0.7x) & 2.7 (0.9x) & 0.7 (0.2x) & 5.5 (1.8x)             \\
Frames (K)          & 122  & 75 (0.6x)& 98 (0.8x) & 25 (0.20) & 198 (1.6x)    \\
Agent per Frame (Average)	    & 5    & 16 (3.2x) & 15 (3.0) & 28 (5.6) & 22 (4.4x)     \\
Agent Tubes (K)     & 7    & 16 (2.3x) &5 (0.7x) & 33 (4.7x) &54 (7.7x)   \\
Agent Labels (K)    & 559  & 1,082 (1.9x) &  2,723 (4.9x) & 338 (0.6x) & 4,143 (7.4x) \\
Action Labels (K)   & 641  & 1,146 (1.8x) & 2,918 (4.5x) & 361 (0.6x) & 4,433 (6.9x)  \\
Location Labels (K) & 498  & 1,130 (2.3x) & 2,804 (5.6x) & 361 (0.7x) & 4,296 (8.6x)   \\
Event Labels (K)    & 548  & 880 (1.6x) & 2,041 (3.7x) & 269 (0.5x) & 3,191 (5.8x) \\ \bottomrule
% Weather Conditions  & 3    & 6          & 6      \\ 
% Storage (Gb)        & 40   & 70         & 110    \\ 
\end{tabular}
}
% \end{table}
% \begin{wraptable}{r}{0.48\linewidth}
% \vspace{-3mm}
\end{table*}

% {\color{red}
\subsection{Human Annotation Process}

All annotations in ROAD-Waymo were produced by trained professional annotators with prior experience in video-based autonomous driving datasets. Annotators underwent an onboarding phase consisting of detailed written guidelines, example-driven training sessions, and pilot annotation rounds reviewed by domain experts.

Annotations follow the \{Agent, Action, Location\} protocol introduced in ROAD. For each agent tube, labels were updated whenever a meaningful semantic change occurred, rather than at a fixed temporal interval. Ambiguous cases (e.g., overlapping actions, partial occlusions, or visually similar agent categories such as bicycles and motorcycles) were flagged by annotators and resolved through expert review.

Quality control was conducted in multiple stages. First, a subset of sequences was double-annotated to identify systematic ambiguities. Second, all annotations were passed through the logical verification framework described in Section~\ref{sec:annotation_process}. Finally, all automatically flagged violations and edge cases were manually reviewed and corrected. While inter-annotator agreement is difficult to define for dense, multi-label spatiotemporal annotations, consensus-based expert adjudication was used for all disputed cases.
% }

\subsection{Style of annotation} %Carried Forward from \road} %% REZA

% \sk{Edit finalise}
% \textbf{Summary of what was annotated in the \road dataset, what types of actors, how many classes etc}
% \AB{@Reza: Some of the below should be moved to the \road dataset and Baseline section - i.e. the description of the existing \road dataset.
% This section below should contain a review of what is the same in \roadpp as \road}

%In accordance with 
Similarly to
the \road dataset~\citep{singh2022road}, three types of labels are used: \textit{i) agents}, \eg cars, cyclists or pedestrians; \textit{ii) actions}, \eg moving, turning or crossing; and \textit{iii) locations}, \eg  junction, incoming-lane or pavement. A complete list of all three types of labels is provided in Appendix A.1 Labels, Table 1.
All of the images are recorded from the (front) camera installed upon the vehicle and they are labelled from the ego-vehicle (\ie AV) point of view (\eg we indicate with ``Incoming lane'' the lane with incoming vehicle from the point of view of the ego vehicle). 
%Among different agents, actions and locations in road environment, 
%Similarly to \road, 
In \roadwaymo we provide annotation for 12 agent classes, 30 action classes and 16 location classes. Different combinations of classes from these three label types yield a considerable number of \textit{road events}. For instance, the road event ``\textit{Car is Moving in Vehicle lane}", is composed of a triplet containing an agent (Car) performing an action (Moving) in a location (Vehicle-lane).
A single agent may perform multiple actions, \eg ``\textit{a car is indicating left while turning left}" and be associated with multiple locations, \eg ``\textit{a pedestrian is on the left pavement and at the bus stop}''. 

\subsection{Annotation process}
\label{sec:annotation_process}

\paragraph{Tracks and Boxes.} In \roadwaymo, all agents inherit bounding box and track annotation from the Waymo-Open dataset~\citep{sun2020scalability}. As bounding box and track annotation were not originally provided by Waymo for traffic lights, we added them using a 
YOLOv6 detector~\citep{li2022yolov6} for initialisation, to later manually filter or adjust the annotation.

\paragraph{Labelling.} Labelling was performed manually for all types of labels. %following the structure of the existing \road dataset. 
Annotators labelled the agent type once per track, then propagated the labels in time for the entire track. Action
and location labels were annotated at the start of the track, while class labels were added to or removed from frames whenever meaningful variations occurred.

% {\color{red} 
\paragraph{Temporal Density of Annotations.}
Action and location labels are temporally sparse and updated only when semantic changes occur. Across the dataset, the average action segment duration is on the order of a few seconds, with a median update interval of 32 frames. This design balances annotation cost with semantic fidelity and reflects the natural temporal persistence of driving behaviors.
% }

\paragraph{Annotation Verification.} %inspired by the machine learning operation (MLOps) field, 
Inspired by the core principles of software engineering, before starting with the annotation process we elicited a set of contextual ``commonsense" requirements written in propositional logic and expressing facts that are always true in practice, such as ``a traffic light cannot be red and green at the same time" or ``an agent cannot move away and towards the vehicle at the same time''.\footnote{To elicit the requirements, we adapted the propositional logic requirements from ROAD-R~\citep{giunchiglia2022road}, resulting in a set of 251 requirements.} This allowed us to easily automatically check if {\sl all} annotations were compliant with the requirements.
%using a SAT solver\footnote{We used the {\sc MiniSAT} solver. Link: http://minisat.se.}. 
All found violations were reported, and passed back to the annotators for further revision.
This process was repeated (four times) until no more violations were found. Thanks to our approach we were able to highlight and rectify 6,276 tubes annotations. A more detailed explanation of the annotation verification pipeline can be found in Appendix A.3.
%See supplementary material for the list of different requirements and a visualisation of the annotation pipeline. 
\roadwaymo is the first dataset whose annotations are guaranteed to be compliant with a set of requirements, and it thus shares the same spirit of  MLOps works such as \citep{gebru_datasheets,mitchell2018_cards,manifesto}, which advocate for a more structured approach to machine learning model development.

\subsection{Statistics}

\begin{figure}[t]
    \centering
    \includegraphics[trim={0.15cm 0.15cm 0.15cm 0.15cm},clip,width=0.90\linewidth]{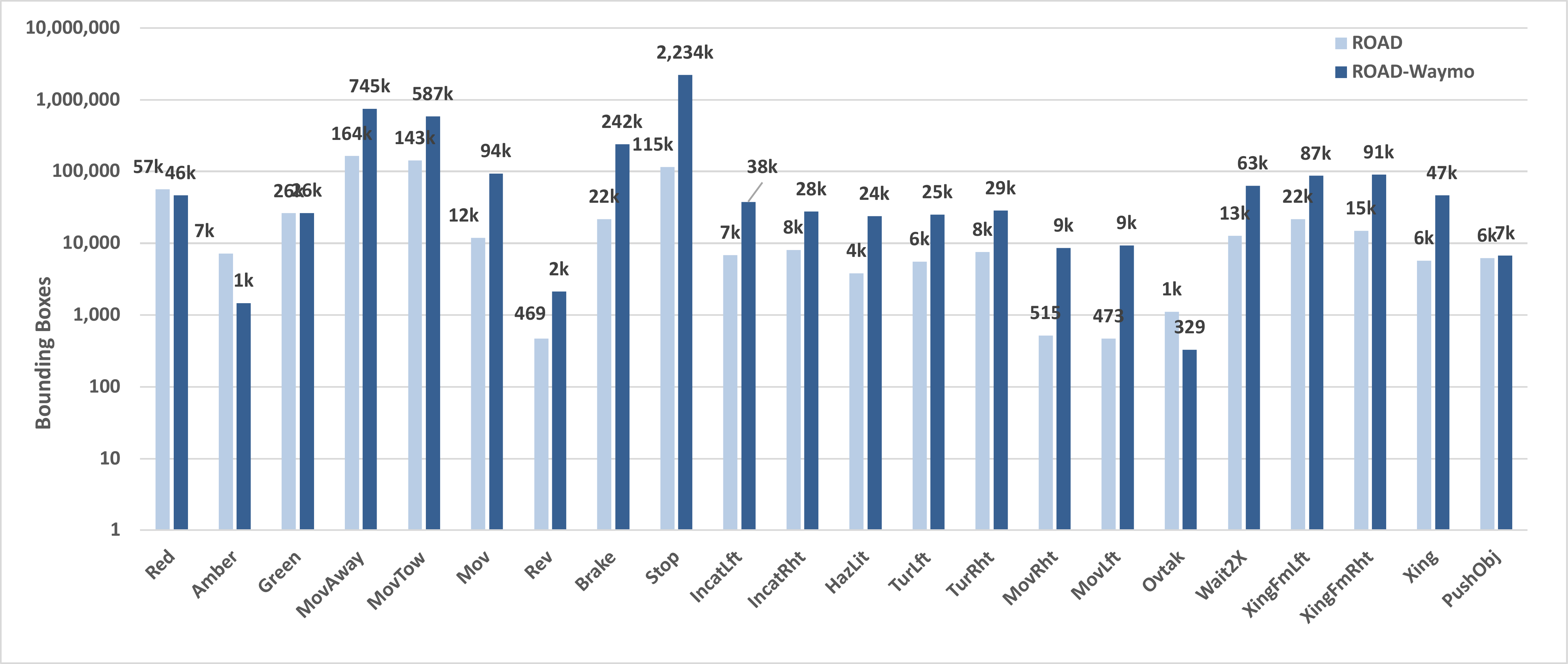}
    \caption{Action labels and count in logarithmic scale. The full description of each abbreviation is provided in Appendix A.1, Table 1.}
    % {\color{red} the abbreviations are not obvious, there should be explained in the supplementary and referenced here; OR add full descriptions here}
    \label{fig:action_stat}
    \vspace{-4mm}
\end{figure}

\begin{figure*}[t]
    \centering
    \includegraphics[trim={0.15cm 0.15cm 0.15cm 0.15cm},clip,width=0.90\linewidth]{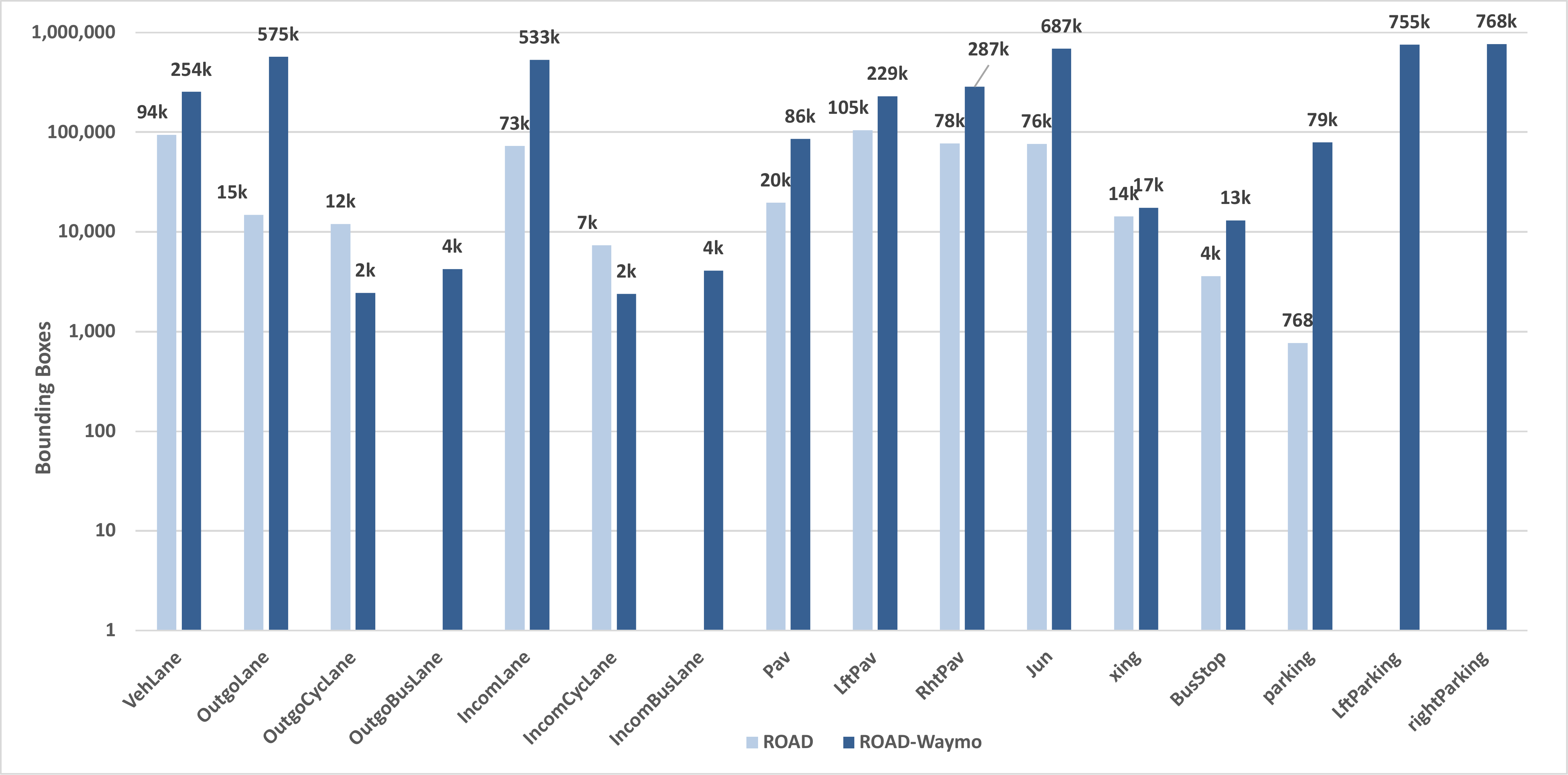}
    \caption{Location labels and count in logarithmic scale. The full description of each abbreviation is provided in Appendix A.1, Table 1.}
    % {\color{red} there is an apparent discrepancy on the right, care to explain it somewhere in the text? also please refer to supplementary for abbreviations (you can do it once in the text maybe)}
    \label{fig:location_stat}
    % \vspace{-2mm}
\end{figure*}

%In comparison to the \road dataset, \roadwaymo surpasses it in three key aspects. 
\roadwaymo is considerably larger than the existing \road dataset (Table~\ref{tab:rpp_stats}).
% this purpose. It %Firstly, in terms of inclusion, \roadwaymo
It encompasses a greater number of agents, actions, and locations, %and in statistical terms, the magnitude of \roadwaymo is significantly larger, 
with tubes and annotations 
\begin{figure}
  \begin{center}
  % \vspace{-0.6cm}
    \includegraphics[width=0.98\linewidth]{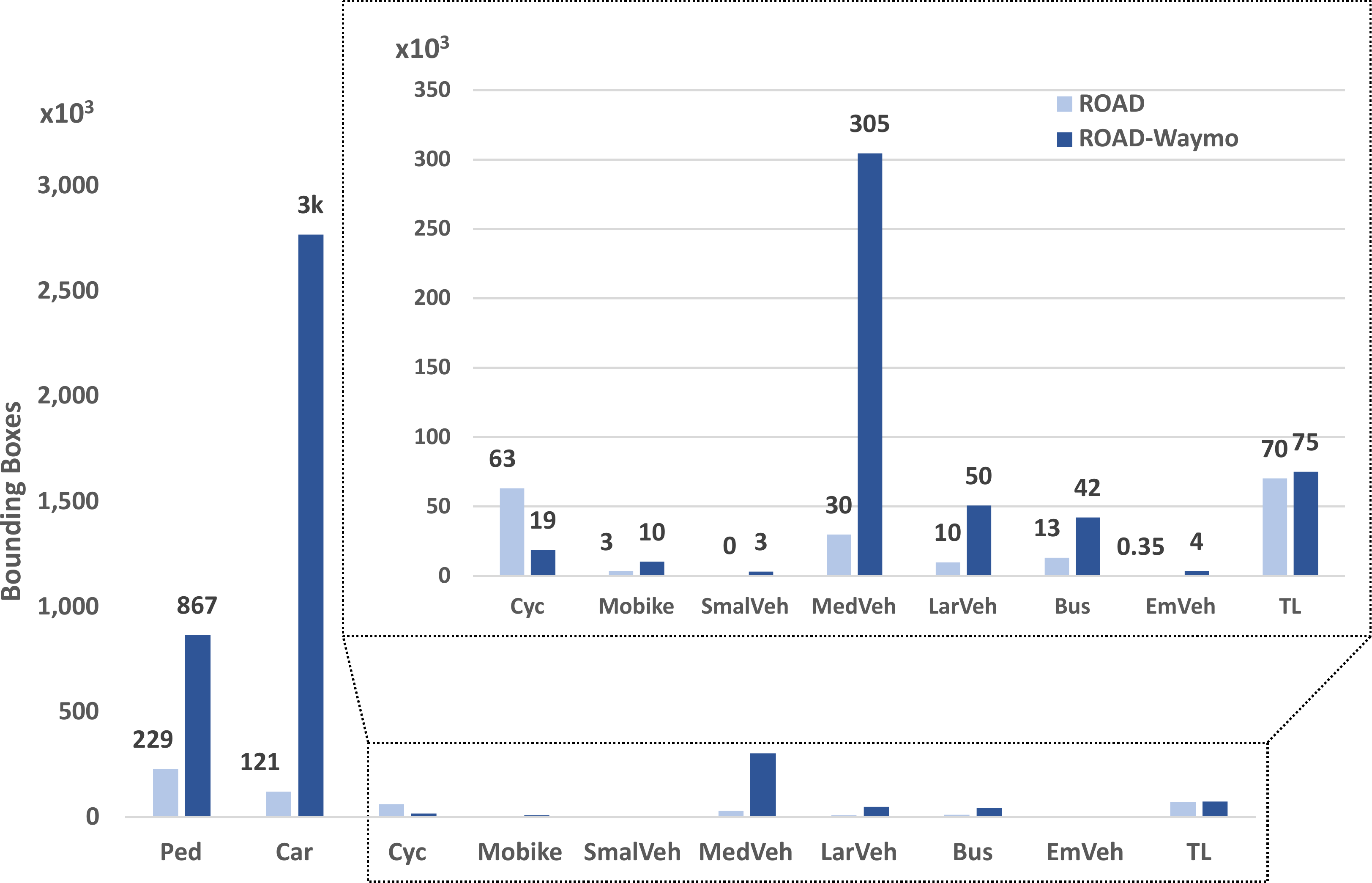}
    % \vspace{-3mm}
    \caption{Agent labels and count in normal scale. \label{fig:agent_zoom} }
      % \vspace{-0.8cm}
\end{center}
\end{figure}
averaging eight times the number. %Lastly, in relation to diversity, 
\roadwaymo exhibits a broader scope, encompassing four distinct (US) cities across six different weather conditions, while \road is confined to a single (UK) city, with half the range of weather conditions. 
% A comprehensive comparison between \road and \roadwaymo 
%, and their combined dataset (\roadpp) 
% is provided in Table \ref{tab:rpp_stats}.
% For a visual representation of the discrepancy between \road and \roadwaymo in terms of agent labels and counts, refer to Figure \ref{fig:agent_stat}. 
As a general theme, \roadwaymo has many more vehicles (\eg 30 times more cars, see Figure~\ref{fig:agent_zoom}) but fewer cyclists. This is just one example of a challenge encountered when shifting domain to another country. \textit{Small-vehicle} and \textit{Emergency-vehicle} had no instances at all in \road, while \roadwaymo contains more than 3K such instances. The presence of emergency situations is arguably extremely important to test the adaptation ability and safety of autonomous vehicles. Similar increments 
%in the number of instances of other vehicle categories 
can be observed for other vehicle categories. Detailed visual comparisons between \road and \roadwaymo in terms of the number of Agents, Actions, and locations are shown in Figures \ref{fig:agent_zoom}, \ref{fig:action_stat} and \ref{fig:location_stat}, respectively. Note that Figures \ref{fig:action_stat} and \ref{fig:location_stat} illustrate the comparison in logarithmic scale.

\section{Baselines}

\subsection{Models \& Architectures}
\textbf{3D-RetinaNet}~\citep{singh2022road} is designed for event detection in autonomous driving. Built upon the popular RetinaNet architecture (originally for object detection in 2D images), it leverages both spatial and temporal information to process video sequences - enabling detection and classification of actions. To process the input sequence of frames, it utilises a variety of 3D backbones---including Inflated 3D (I3D) \citep{carreira2017quo} and SlowFast \citep{feichtenhofer2019slowfast}---to capture hierarchical and multi-scale representations, which are used within two modules to capture and predict motion and temporal dynamics. Here, we update the 3D-RetinaNet to be trained on \roadwaymo and/or combined (\road + \roadwaymo) datasets.
\textbf{I3D backbone.} Inflated 3D (I3D) \citep{carreira2017quo} is a spatiotemporal backbone that extends conventional 2D convolutional networks to video by “inflating” filters and pooling operations into the temporal dimension, allowing the model to jointly learn appearance and motion cues from short frame sequences. This design makes I3D well-suited to capturing fine-grained temporal patterns (e.g., subtle changes in pose, interactions, or object dynamics) while still benefiting from strong spatial feature extraction at the frame level. In our setup, I3D is used as the 3D feature extractor within 3D-RetinaNet, producing temporally-aware feature maps that can be consumed by the detection heads to localise and classify events/actions across time.

\textbf{SlowFast backbone.} SlowFast \citep{feichtenhofer2019slowfast} is a two-pathway video architecture designed to balance semantic understanding and motion sensitivity: a “Slow” pathway processes frames at a lower temporal rate to capture rich spatial semantics, while a “Fast” pathway runs at a higher frame rate with a lightweight channel capacity to better model rapid motion and temporal changes. Features from both pathways are fused, yielding multi-scale representations that are often stronger for actions/events where both context (scene layout, agent identity) and dynamics (changes over time) matter. In our experiments, SlowFast serves as an alternative 3D backbone inside 3D-RetinaNet, providing complementary motion-centric representations that can improve event/action detection performance in driving videos, particularly for interactions that unfold over multiple frames.

\textbf{YOLOv8} 
~\citep{Jocher_YOLO_by_Ultralytics_2023} represents a highly advanced and top-of-the-line model that takes the accomplishments of previous YOLO iterations to new heights. It incorporates novel elements and enhancements that significantly enhance performance and adaptability, with its focus on speed, precision, and user-friendliness. 
%YOLOv8 emerges as an exceptional option for various tasks, including object detection and tracking, instance segmentation, image classification, and pose estimation. 
Here, we train the YOLOv8 model for object (road agent) detection tasks.
% We will show remarkbley good resuts using such an ef
% {\color{red}Any implementation/parameter setting details to mention, for this or the other baselines?}

\textbf{Domain Adaptation Baseline}. %% Izzeddin
\label{sec:dom_adapt_baseline}
%Consistent with the established practice in domain adaptation research, we employed 
Reverse Gradient (RevGrad)  \citep{ganin2015unsupervised} is a popular baseline approach to domain adaptation. This works by minimising the discrepancy between the source and target domains, by iteratively updating the model parameters such that the gradient of the adaptation loss with respect to the feature extractor is reversed during backpropagation - effectively aligning the feature distributions of the two domains. This encourages the model to learn domain-invariant representations that can generalise to the target domain, while retaining task-specific information from the source domain. 
In our tests, this model is trained on both domains UK \road and US \roadwaymo
% {\color{red} which domains? countries? cities? which experiments are we actually running?}
(source with labels, and target without labels), and then tested on the target domain (see Sec. \ref{sec:road++}).

\textbf{Neuro-Symbolic Baseline}. %%Ele % Mihaela:shortened this section,  check it reads well with the rest of the baselines and the experimental analysis sectio
Commonsense, propositional logic requirements can be used for more than just checking annotation validity (as described above). They also allow \roadwaymo to be utilised as a benchmark for neuro-symbolic models  able to leverage the background knowledge expressed by these requirements to get better performance. To this end, we provide a neuro-symbolic baseline 
%in which we 
integrating the requirements in the loss,  
%we also utilised them at training time by integrating them into a new loss term added to the usual loss objective,
following a similar procedure as in \road-R~\citep{giunchiglia2022road}. This integration has been shown  to improve the models' performance in  different works (\eg \cite{diligenti2017icmla,diligenti2017semantic,xu2018semanticloss,giunchiglia2021,serafini2022ltn,stoian2023tnorms}).
% To this end, we equipped \roadwaymo with the set of 272 constraints we used for the automatic annotation checks, capturing background knowledge. 
% such as ``an agent cannot move away and towards the self-driving vehicle at the same time'' or ``a pedestrian cannot wait to cross and cross at the same time'', and then we incorporated these rules into a logic-based loss.
% The new loss is simply added as a regularization term into the usual loss objective of the models and measures the degree of constraint satisfaction w.r.t. the neural predictions by mapping Boolean logical constraints into the real domain and Boolean logic operators into differentiable algebraic operations, called t-norms \citep{metcalfe2005}.
% The new loss was simply added as a regularization term into the usual loss objective of the models and
The new loss measures the degree of constraint violation with respect to the neural predictions by mapping Boolean logic operators into differentiable algebraic operations, called t-norms \citep{metcalfe2005}.
Ultimately, this enhances neural network prediction by allowing the models access to background knowledge during training.
% When computing this new loss term, we considered two variants based on two different t-norms\citep{metcalfe2005}: G\"{o}del and \L{}ukasiewicz.
% Ultimately, this provides a way of enhancing the decision making of a neural network by allowing it access to background knowledge during training and has been previously shown  \citep{diligenti2017icmla,donadello2017ltn_sii,xu2018semanticloss,diligenti2017semantic,fischer2019dl2,ahmed2022a,giunchiglia2022road} to improve the models' performance. 
% Ultimately, it provides a way of enhancing the decision making of a neural network by allowing it access to background knowledge during training and has been previously shown  \citep{diligenti2017icmla,donadello2017ltn_sii,xu2018semanticloss,diligenti2017semantic,fischer2019dl2,ahmed2022a,giunchiglia2022road} to improve the models' performance. 
% Additionally, similar neuro-symbolic approaches have been previously shown  \citep{diligenti2017icmla,donadello2017ltn_sii,xu2018semanticloss,diligenti2017semantic,fischer2019dl2,ahmed2022a,giunchiglia2022road} to improve the models' performance. 
% {\color{red}
\paragraph{Choice of Baselines.}
We emphasise that the selected baselines are not intended to represent the state of the art in video understanding, but rather to provide strong, well-understood reference points that are widely used and reproducible. Recent advances such as video transformers, masked autoencoders, and contrastive or source-free domain adaptation methods (e.g., VideoMAE~\cite{wang2023videomaev2scalingvideo}, ActionFormer~\cite{zhang2022actionformerlocalizingmomentsactions}, STAR~\cite{gritsenko2024end} ) are promising directions for future evaluation on ROAD-Waymo and ROAD++.
% }

\subsection{Implementation Details}

\textbf{Tasks}. We follow the experimental setup of \road \citep{singh2022road}, and evaluate on \roadwaymo approaches over agent, action, location, duplex (joint label of agent+action) and event detection tasks, along with the temporal segmentation of the action performed by the AV itself. 
%AV-action segmentation task on the newly introduced \roadwaymo dataset and compare them with the smaller \road dataset.
In addition, we introduce a set of experiments within the \roadpp framework, where we study unsupervised domain adaption from one source dataset to a target dataset on the aforementioned tasks. In our tests, either \road or \roadwaymo play the role of the source dataset, with the other acting as target. 
%then the target dataset would be the other one.}

\textbf{Metrics}. Similarly to \cite{singh2022road}, in our experiments results are evaluated for both \textit{frame-level} (bounding box) detection and \textit{video-level} (tube) detection. 
As evaluation metrics %used for frame-level bounding box and tube detection are 
we use frame mean Average Precision (f-mAP) and video mean Average Precision (v-mAP), respectively, commonly used in action detection \citep{singh2017online,kalogeiton2017action,saha2016deep}. The detection threshold for Intersection over Union (IoU) is set to 0.5 for f-mAP, indicating a 50\% overlap between predicted and actual bounding boxes. To compute the v-mAP, instead, the results are evaluated at 0.2 IoU threshold, \ie 20\% overlap,
%on two IoU thresholds, namely, 0.2 (20\% overlap) and 0.5 (50\% overlap) 
as detecting tubes is more challenging than detecting single bounding boxes. %\AB{should that say 20\% overlap?}

\textbf{Standard supervised training.}
We train 3D-RetinaNet~\citep{singh2022road} with I3D~\citep{carreira2017quo} and SlowFast~\citep{feichtenhofer2019slowfast}. Here, we update the 3D-RetinaNet to be trained on ROAD-Waymo and/or combined (ROAD + ROAD-Waymo) datasets.

\textbf{Unsupervised domain adaptation (RevGrad).}
Reverse Gradient (RevGrad)  \citep{ganin2015unsupervised} is a popular baseline approach to domain adaptation. This works by minimising the discrepancy between the source and target domains, by iteratively updating the model parameters such that the gradient of the adaptation loss with respect to the feature extractor is reversed during backpropagation - effectively aligning the feature distributions of the two domains. This encourages the model to learn domain-invariant representations that can generalise to the target domain, while retaining task-specific information from the source domain. 
In our tests, this model is trained on both domains UK \road and US \roadwaymo (source with labels, and target without labels), and then tested on the target domain (see Sec. \ref{sec:road++}).
%%%%%%%%%%%%%%%%%%%%%%%%%%%%%

\textbf{Neuro-Symbolic training (constraints in the loss).}
Commonsense, propositional logic requirements can be used for more than just checking annotation validity (as described above). They also allow \roadwaymo to be utilised as a benchmark for neuro-symbolic models  able to leverage the background knowledge expressed by these requirements to get better performance. To this end, we provide a neuro-symbolic baseline 
%in which we 
integrating the requirements in the loss,  
%we also utilised them at training time by integrating them into a new loss term added to the usual loss objective,
following a similar procedure as in \road-R~\citep{giunchiglia2022road}. This integration has been shown  to improve the models' performance in  different works (\eg \cite{diligenti2017icmla,diligenti2017semantic,xu2018semanticloss,giunchiglia2021,serafini2022ltn,stoian2023tnorms}).

The new loss measures the degree of constraint violation with respect to the neural predictions by mapping Boolean logic operators into differentiable algebraic operations, called t-norms \citep{metcalfe2005}.
Ultimately, this enhances neural network prediction by allowing the models access to background knowledge during training.
%%%%%%%%%%%%%%%%%%%%%%%%%%%%%%%%%%%%%%%

\textbf{Model Training.}
To train these models, we fix the input sequence to have length equal to 8 frames with an image size of 600$\times$840. 
The backbone networks are initialised using weights from Kinetics~\citep{kay2017kinetics} pretraining. We employ an SGD optimizer with a step-learning rate, initially set to \textit{0.01$\times$number of GPUs} to then decrease by a factor of 10 after 18 and 25 epochs, totaling 30 epochs overall. 

\textbf{Testing}. Although we train for 8 frame sequences, we can test the same network for longer input sequences as well. In our experiments we found that testing with 32 frames as input yielded slightly better results.

\textbf{Hardware.} All the experiments are performed using a max of GPU servers including 8 × Nvidia-A30 with 24GB onboard memory (VRAM), 8 × Nvidia-RTX 6000 with 24GB VRAM, and 3 × Nvidia-A100 with 80GB VRAM. To ensure a fair comparison of the results, consistent hyperparameters, as specified in the main paper, were maintained across all the servers.
%%%%%%%%%%%%%%%%%%%%%%%%%%%%%%%%%%%%%%%%%%%%%%%%%%%%
% \subsection{Dataset Splits}
% %
% %%%%%%%%%%%%%%%%%%%%%%%%%%%%%%%%%%%%%%%%%%%%%%%%%%%%
% \section{Results \& Analysis}
% %
% %%%%%%%%%%%%%%%%%%%%%%%%%%%%%%%%%%%%%%%%%%%%%%%%%%%%
% \subsection{Standard Detection Benchmarks}
% %
% %
% %%%%%%%%%%%%%%%%%%%%%%%%%%%%%%%%%%%%%%%%%%%%%%%%%%%%
% \subsection{Domain Adaptation (ROAD++)}
% %
% %
% %%%%%%%%%%%%%%%%%%%%%%%%%%%%%%%%%%%%%%%%%%%%%%%%%%%%
% \subsection{Neuro-Symbolic Learning}
%

%%%%%%%%%%%%%%%%%%%%%%%%%%%%%%%%%%%%%%%%%%%%%%%%%%%%

\subsection{Experimental Setup}

% {\color{red}Please describe tasks and metrics.}
%% GUR
\textbf{Tasks}. We follow the experimental setup of \road \citep{singh2022road}, and evaluate on \roadwaymo approaches over agent, action, location, duplex (joint label of agent+action) and event detection tasks, along with the temporal segmentation of the action performed by the AV itself. 
%AV-action segmentation task on the newly introduced \roadwaymo dataset and compare them with the smaller \road dataset.
In addition, we introduce a set of experiments within the \roadpp framework, where we study unsupervised domain adaption from one source dataset to a target dataset on the aforementioned tasks. In our tests, either \road or \roadwaymo play the role of the source dataset, with the other acting as target. 
%then the target dataset would be the other one.}

\textbf{Metrics}. Similarly to \cite{singh2022road}, in our experiments results are evaluated for both \textit{frame-level} (bounding box) detection and \textit{video-level} (tube) detection. 
As evaluation metrics %used for frame-level bounding box and tube detection are 
we use frame mean Average Precision (f-mAP) and video mean Average Precision (v-mAP), respectively, commonly used in action detection \citep{singh2017online,kalogeiton2017action,saha2016deep}. The detection threshold for Intersection over Union (IoU) is set to 0.5 for f-mAP, indicating a 50\% overlap between predicted and actual bounding boxes. To compute the v-mAP, instead, the results are evaluated at 0.2 IoU threshold, \ie 20\% overlap,
%on two IoU thresholds, namely, 0.2 (20\% overlap) and 0.5 (50\% overlap) 
as detecting tubes is more challenging than detecting single bounding boxes. %\AB{should that say 20\% overlap?}
% {\color{red}
\paragraph{Dataset Splits.}
ROAD-Waymo strictly follows the official training, validation, and test splits defined by the Waymo Open Dataset to ensure full alignment with the original benchmark and its associated challenges. By adopting the exact same splits, we preserve consistency with existing Waymo-based evaluations and enable fair comparison with prior and future methods developed on the Waymo ecosystem. In accordance with Waymo’s evaluation protocol, annotations for the test set are not publicly released. Instead, test-set evaluation is performed through the official Waymo challenge server, preventing overfitting and ensuring the integrity of benchmark results.
% }

\begin{table}
    \centering
    \setlength{\tabcolsep}{4pt}
    \caption{\roadwaymo frame-level results (f-mAP$\%$) Val/Test. }
    % \vspace{-3mm}
    {\footnotesize
    \begin{tabular}{lccccc}
    \toprule
    Model &  Agents & Actions & Locations & Duplexes & {Events} \\ \midrule
    I3D-08     & 15.7/16.7 & 12.3/13.9 & \textbf{12.4}/\textbf{13.2} & 10.5/10.1 &  6.2/5.3\\
    I3D-32     &  15.4/15.2 & 12.2/13.0 & 12.3/12.8 & 10.4/9.4 & 6.3/5.6\\
    SlowFast-08   &  16.1/15.3 & \textbf{13.0}/\textbf{14.0} & 11.9/12.4 & \textbf{10.7}/\textbf{10.2} & 4.7/5.7\\
    SlowFast-32      & 16.0/15.0 & \textbf{13.0}/13.8 & 11.9/12.2 & \textbf{10.7}/\textbf{10.2} & \textbf{6.8}/\textbf{5.8}\\
    YOLOv8      & \textbf{38.1}/\textbf{31.6} & -/- & -/- & -/- & -/-\\
   \bottomrule
    \end{tabular}
    }       
    \label{tab:Roadwaymoframe-avg-results} 
    \vspace{-2mm}
\end{table} 

\textbf{Model training}. We train 3D-RetinaNet~\citep{singh2022road} with I3D~\citep{carreira2017quo} and SlowFast~\citep{feichtenhofer2019slowfast}. To train these models, we fix the input sequence to have length equal to 8 frames with an image size of 600$\times$840. 
%The selection of $T$ and image size is 
% This choice is induced by GPU memory constraints and makes the baseline compatible with both \road and \roadwaymo. During testing, however, we extend the convolutional 3D-RetinaNet to handle sequences of 32 frames, enabling streaming deployment. {\color{red}Wanna make reference to continual learning/detection papers here?}
The backbone networks are initialised using weights from Kinetics~\citep{kay2017kinetics} pretraining. We employ an SGD optimizer with a step-learning rate, initially set to \textit{0.01$\times$number of GPUs} to then decrease by a factor of 10 after 18 and 25 epochs, totaling 30 epochs overall. 

\textbf{Testing}. Although we train for 8 frame sequences, we can test the same network for longer input sequences as well. In our experiments we found that testing with 32 frames as input yielded slightly better results.

% }
\section{Experiments Results}
%of previous YOLO iterations to new heights. It incorporates novel elements and enhancements that significantly enhance performance and adaptability, with its focus on speed, precision, and user-friendliness. Here, we train the YOLOv8 model for object (road agent) detection tasks.
%
%%%%%%%%%%%%%%%%%%%%%%%%%%%%%%%%%%%%%%%%%%%%%%%%%%%%

\color{black}
\subsection{\roadwaymo Results}

% {\color{red}My main problem with this section is the limited critical reflection - what do those results mean? What is their significance for the computer vision and AD community? Also, on agent/action/event detection you make no comparison with the baseline results on ROAD - how much more challenging is ROAD-Waymo? I'd like to see more discussion}
    
% \begin{wraptable}{r}{0.64\linewidth}
%     \centering
%     \setlength{\tabcolsep}{4pt}
%     \caption{\roadwaymo Frame-level results (mAP $\%$)  }
%     \vspace{-2mm}
%     {\footnotesize
%     \scalebox{0.85}{
%     \begin{tabular}{lcccccc}
%     \toprule
%     Model & {Agentness}  & Agents & Actions & Locations & Duplexes & {Events} \\ \midrule
%     I3D-08     & 31.8/36.3 & 15.7/16.7 & 12.3/13.9 & 12.4/13.2 & 10.5/10.1 &  6.2/5.3\\
%     I3D-32     & 31.5/32.6 & 15.4/15.2 & 12.2/13.0 & 12.3/12.8 & 10.4/9.4 & 6.3/5.6\\
%     SlowFast-08     & 29.9/31.9 &  16.1/15.3 & 13.0/14.0 & 11.9/12.4 & 10.7/10.2 & 4.7/5.7\\
%     SlowFast-32     & 29.8/31.8 & 16.0/15.0 & 13.0/13.8 & 11.9/12.2 & 10.7/10.2 & 6.8/5.8\\
%     YOLOv8     & -/- & 38.1/31.6 & -/- & -/- & -/- & -/-\\
%    \bottomrule
%        \vspace{-8mm}
%     \end{tabular}
%     }
%     }
%     \label{tab:Roadwaymoframe-avg-results} 
% \end{wraptable}

\textbf{Frame-level Results.} Table \ref{tab:Roadwaymoframe-avg-results} presents the frame-level results for all five tasks %{\color{red} I don't believe we explained them - certainly not agentness and duplex detection} 
using f-mAP with an IoU threshold of 0.5. 
%In Table \ref{tab:Roadwaymoframe-avg-results}, the 
Results are reported for both validation and test sets {separated by a forward slash (/)}. Rows in the Table represent combinations of feature extraction backbones (I3D and SlowFast) for the 3D-RetinaNet baseline. 
%It can be seen from the Table that 
SlowFast with a test sequence length of 8 achieves the best performance for actions and duplex while, for events, SlowFast-32 reports the best performance. 
% {\color{red} text description completely wrong, does not match the table at all}

We also report the 
%results of the recent state-of-the-art object detection model 
performance of state-of-the-art
YOLOv8~\citep{Jocher_YOLO_by_Ultralytics_2023} for agent (object) detection, which indeed achieves the best performance for the task. The YOLOv8 model is trained and tested on a single frame rather than sequences, which is the main reason 
%for training 
why we trialled it
for agents only. 
Its
%YOLOv8 show 
remarkable performance 
%on the agent detection task 
%because it make use of many tricks 
is likely due to its use of several expedients such as: multi-scale training; mosaic data augmentation \etc. 

\begin{table}

  \centering
  \setlength{\tabcolsep}{4pt}
  \caption{\roadwaymo video-level results (v-mAP$\%$) Val/Test. }
  % \vspace{-3mm}
  {\footnotesize
  \scalebox{0.99}{
  \begin{tabular}{lccccc}
  \toprule
  Model & Agents & Actions & Locations & Duplexes & {Events}\\ \midrule
  \midrule
  % \multicolumn{6}{l}{Detection threshold $\delta = 0.2$} \\
  % \midrule

  I3D-08    &  4.8/5.2 & 4.5/\textbf{4.6 }&  4.2/6.1 & 6.2/5.7 &  4.2/\textbf{4.3}\\
  I3D-32    &  5.5/4.6 & \textbf{4.6}/4.3 & \textbf{4.4}/\textbf{6.2} & \textbf{6.2}/5.9 & \textbf{4.7}/4.2\\
  SlowFast-08    &  \textbf{6.5}/\textbf{5.5} & 4.2/\textbf{4.6} & 4.2/5.4 & 5.9/\textbf{6.3} & \textbf{4.7}/ 4.1\\
  SlowFast-32    &  6.4/5.1 & 4.1/4.4 & 4.3/4.7 & 5.5/6.1 & 4.3/4.0\\
  
  % \midrule
  % \multicolumn{6}{l}{Detection threshold $\delta = 0.5$} \\ 
  % \midrule
  % I3D-08    & 1.9/1.2 & \textbf{1.4}/1.3 & 1.0/1.4 & \textbf{2.8}/2.3 & 1.8/\textbf{1.7}\\
  % I3D-32    & 1.6/\textbf{1.3} & 1.2/\textbf{1.4} & \textbf{1.1}/\textbf{2.1} & 2.5/\textbf{2.5} & 1.8/1.6\\
  % SlowFast-08    &  \textbf{2.6}/ 0.8 & 0.9/0.9 & 0.9/1.0 & 2.0/ 1.5 & \textbf{2.0}/1.1\\
  % SlowFast-32    &  \textbf{2.6}/0.9 & 0.8/0.9 & 0.9/0.9 & 1.6/1.6 & 1.4/1.1\\
  \bottomrule
  \end{tabular}
  }
  }
    \vspace{-4mm}
  \label{tab:Road-waymo-video-avg-results} 
\end{table}

\textbf{Video-level Results.} Table \ref{tab:Road-waymo-video-avg-results} presents video-level results in terms of v-mAP. Results are reported for both 3D-RetinaNet backbones (I3D and SlowFast) and sequence lengths values (8 and 32 frames). V-mAP is reported for IoU thresholds $\delta$ of 0.2. 
%With a lower value of $\delta$ = 0.2 
SlowFast-08 reports the best performance for agents, actions, and duplex, while for locations and events 
%the best methods are 
I3D-32 and I3D-08
top the charts.
%, respectively. 
% However, for a higher value of overlap (50\%), I3D-32 achieves the best scores for all label types (except events, for which the best performer is I3D-08). {\color{red} here again results look more mixed than what described} Overall, the lower video-level results obtained from 3D-RetinaNet for all label types highlight the significant challenges posed by the \roadwaymo dataset for video-level detection.

\begin{table}[h!]

  \centering
  \setlength{\tabcolsep}{4pt}
  \caption{Temporal AV-action detection results% (frame mAP$\%$).
  }
  % \vspace{-3mm}
  {\footnotesize
  \scalebox{0.99}{
  \begin{tabular}{lccc}
  \toprule
  & No instances  & \multicolumn{2}{c}{f-mAP@$0.5$ (\%) %/Video-mAP@$0.2$
  } \\ 
  \midrule
  Model & \multicolumn{1}{c}{} & I3D & SlowFast
  \\ 
  \midrule 
  Eval subset & train / val / test & val / test  & val / test \\ 
  % \midrule
  % \multicolumn{5}{l}{Duplex results} \\ 
 \midrule
  Move                   & 88k / 29k / 30k    & 98.10 / 98.36  & \textbf{98.11} / \textbf{98.46}    \\ 
  Stop                   & 21k / 7.3k / 6.6k    & 94.78 / 95.89  & \textbf{97.57} / \textbf{97.68}    \\ 
  Turn-right             & 1.8k / 0.8k / 1k      & 48.94 / 41.89  & \textbf{59.30} / \textbf{48.40 } \\ 
  Turn-left              & 2.2k / 0.4k / 0.3k      & 31.98 / 47.24 & \textbf{47.38} / \textbf{55.79}  \\ 
  Move-left              & 0.6k / 0.1k / 0.3k       & \textbf{10.43} / \textbf{6.06}  & 6.74 / 4.86 \\ 
  Move-right             & 0.8k / 0.3k / 0.3k       & 3.65 / 14.58  & \textbf{7.20} / \textbf{25.25}   \\ 
  Total/Mean             & 115k / 38k / 39k   & 47.99 / 50.67  & \textbf{52.72} / \textbf{55.08}     \\ 
 \bottomrule
  \end{tabular}
  }
  }
\vspace{-2mm}
\label{tab:classwise-av-actions} 

\end{table}

\textbf{Results of AV-Action Segmentation.} In addition to frame and tube detection, we also analyse the use of 3D-RetinaNet to temporally segment AV-action classes (Table \ref{tab:classwise-av-actions}). Results are reported for I3D and SlowFast backbones over both validation (Val) and test sets. The number of instances for each category is also provided---highlighting the complexity and challenge of recognising these categories in long-tail problems. The performance for the ``AV-move" and ``AV-stop" classes is high, likely because these two classes are predominantly represented in the dataset. The performance for the ``turning" classes is reasonable, but the outcomes for the bottom two classes are notably poor. It can be seen that SlowFast significantly outperforms I3D as it is more suited to a pure classification problem. %{\color{red} Slowfast wins, right?}
% {\color{red}Why do you use the Val/Test columns here instead of just using a backslash in this Table as in the other tables? -- fixed }

% {\color{red}
\textbf{Error Analysis and Failure Modes.}
We observe that most errors occur in long-tail action classes and short-duration behaviors (e.g., lane-change actions), which are challenging due to class imbalance and subtle visual cues. Common failure modes include partial occlusions, visually similar agent categories, and rapid action transitions. Event-level performance remains low, reflecting the compounded difficulty of accurate agent, action, and temporal localization. These challenges highlight the realism and difficulty of the benchmark.
% }

\subsection{\roadpp framework (UK vs US Roads)}%\AB{@Izz pls check new title} 
\label{sec:road++}
% {\color{red} Please review correctness of this paragraph and precisely clarify the three protocols}
% To address the challenging problem of domain shift between different countries, experiments were performed using both \road and \roadwaymo to provide a baseline. 
\roadwaymo has same annotation style as the original \road dataset, allowing us to study the problem of domain shift between UK and USA roads for AVs. We thus introduce a first-of-its-kind experimental setup for cross-dataset and cross-city 
%evaluation and 
real-to-real
unsupervised domain adaptation, for the agent, action and event detection tasks.

\begin{table*}[t]
\vspace*{-0.3cm}
  \centering
  \setlength{\tabcolsep}{4pt}
  \caption{3D-RetinaNet performance in cross-cities and countries (datasets) training and testing. For every city and label subsets we report f-mAP@0.5 and v-mAP@0.2 in percentages, following the convention f-mAP@0.5/v-mAP@0.2. Best results for each label type are in bold.}
  \vspace{-3mm}
  \footnotesize{
\scalebox{0.8}{
\begin{tabular}{llcccccc}
\toprule
% \multicolumn{1}{c}{} & \multicolumn{1}{c}{} & \multicolumn{6}{c}{f-mAP@0.5 (\%) / v-mAP@0.2 (\%)}\\  \midrule
Train \textbackslash  Test                    &           & Oxford  & Phoenix & San Francisco  & Other & \roadwaymo & \roadpp (UK\&US)  \\ \midrule
\multirow{3}{*}{\road (Oxford)}           & Agent     & 24.12 / 12.92 & 1.37 / 3.41  & 4.99 / 4.25 & 0.43 / 3.46 & 5.25 / 2.86 & 14.68 / 7.88 \\ 
& Action    & \textbf{15.92} / 6.96 & 1.08 / 1.68 & 3.56 / 2.69 & 0.27 / 3.83 & 3.5 / 1.84         & 8.62 / 4.4\\ 
& Location  & \textbf{10.51} / 4.53 & 1.01 / 2.02 & 1.52 / 1.76 & 0.17 / 1.22 & 2.47 / 2.25 & 5.17 / 3.39   \\ \midrule

\multirow{3}{*}{Phoenix}           & Agent     & 6.18 / 6.30 & 4.09 / 2.79 & 5.75 / 3.08 & 0.93 / 4.65 & 10.64 / 2.97 & 8.41 / 4.63 \\ 
& Action    & 1.64 / 1.43 & 3.96 / \textbf{3.97} & 4.39 / 2.46 & 0.97 / 5.54 & 9.19 / 2.98 & 5.41 /	2.20 \\  
& Location  & 1.55 / 1.62  & 3.39 / 3.74 & 4.10 / 3.21 & 0.66 / 4.03 & 7.73 / 3.36 & 4.64 / 2.49 \\  \midrule

\multirow{3}{*}{San Francisco}        & Agent     & 10.94 / 7.73 & 3.09 / 2.14 & 10.35 / 5.19 & 1.14 / 5.54 & 14.42 / 3.96 & 12.68 /	5.84 \\ 
& Action    & 5.72 / 1.99 & 2.83 / 2.97 & 7.80 / 3.60 & 1.21 / 5.89 & 11.71 / 3.33 & 8.71 / 2.66 \\  
& Location  & 3.99 / 2.36 & 2.63 /  4.06 & 7.92 / 4.82 & 0.78 / 5.08 & 11.03 / 4.23 & 7.51 / 3.29 \\  \midrule

\multirow{3}{*}{Other}        & Agent     & 6.36 / 5.46 & 2.06 / 1.39  & 4.27 / 2.70 & 0.71 / 2.99 & 6.82 / 2.07 & 6.59 / 3.76 \\ 
& Action    & 1.55 / 0.92 & 1.48 / 2.24 & 2.69 / 2.49 & 0.62 / 3.76 & 4.63 / 2.18 & 3.09 / 1.55 \\  
& Location  & 2.55 / 1.39 & 1.89 / 2.74 & 3.63 / 3.25 & 0.52  / 4.08 &  5.82 / 2.89 & 4.18 / 2.14 \\  \midrule

\multirow{3}{*}{\roadwaymo}     & Agent     & 11.76 / 7.13 & 4.43 / 3.17 & 9.18 / 4.42 & 1.21 / 6.17  & 15.42 / 5.19      & 13.58 / 6.16\\ 
& Action    & 3.94 / 1.95 & 4.18 / 3.96 & 7.38   / 3.80 & 1.18 / 6.26 &13.0 / \textbf{4.58} & 8.47 / 3.26\\ 
& Location  & 3.98 / 2.40 & 3.92 / 7.09  &  7.42 / 4.40 & 0.94 / 4.31 & \textbf{12.9} / \textbf{6.07} & 8.44 / 4.23   \\ \midrule

\multirow{3}{*}{ROAD++ (UK\&US)}        & Agent     & \textbf{26.00} / \textbf{18.34} & \textbf{4.65} / \textbf{4.34} & \textbf{11.10} / \textbf{6.18} & \textbf{1.23} / \textbf{8.09} & \textbf{16.62} / \textbf{5.71} & \textbf{21.31} / \textbf{12.02}  \\ 
& Action    & 15.82 / \textbf{7.57} & \textbf{4.55} / 3.83 & \textbf{8.88}  / \textbf{4.23} & \textbf{1.26} / \textbf{7.25} & \textbf{13.39} / 4.32 & \textbf{14.6} / \textbf{5.94}  \\ 
& Location  & 8.90 / \textbf{6.07} & 4.28 / \textbf{7.74} & \textbf{8.93} / \textbf{5.23} & \textbf{0.96} / \textbf{5.95}   & 12.57 / 4.22 & \textbf{10.73} / \textbf{5.14} \\ \bottomrule
                                
\end{tabular}
}
}
\label{tab:combine} 
\vspace{-2mm}
% \vskip -0.1cm

\end{table*}

\textbf{Cross dataset and cities evaluation}. We report a baseline for cross-dataset and cross-cities evaluation in Table~\ref{tab:combine}, 
where I3D-based 3D-Retinanet
% (, {\color{red}explain what type, training using what version and tested on what version}) 
is trained upon the train sets of \road, \roadwaymo, and  \roadpp (\ie using together the training sets of both \road and \roadwaymo) separately. Then, each of the models is tested upon the test sets of all three mentioned datasets {(where, again, the ROAD++ test set is the union of the test sets of the two datasets)}. 
Detailed f-mAP and v-mAP results are reported in Table \ref{tab:combine}. %{\color{red} what version of 3D retina net are we using here?} 
The IoU threshold was set to 0.5 for frame-mAP and to 0.2 for video-mAP. 
%The ultimate findings derived from these results can be summarised into three key aspects. 
Three key
findings arise from these results.
Firstly, there is a significant reduction in the reported performance, exceeding a three-fold decrease, when training on \road and subsequently testing on \roadwaymo. Secondly, however, this does not hold when the reverse experiment is conducted. Lastly, it can be also observed the best results (as expected) are achieved when training on \roadpp and testing it on individual test folds of \road and \roadwaymo.

\textbf{Unsupervised domain adaptation (UDA).} In adherence to the established practice 
\begin{figure}
% \vskip -1.4cm
    \centering
    \footnotesize
    \includegraphics[trim={0 0.2cm 0 0.2cm},clip,width=0.98\linewidth]{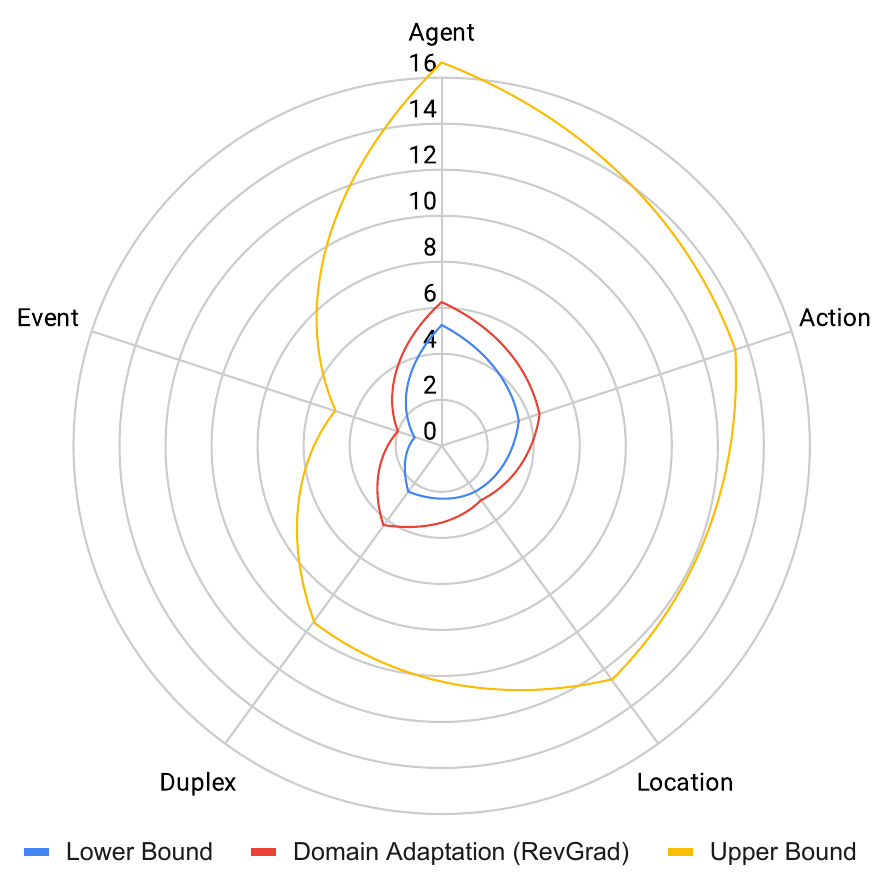}
    \caption{UDA results (f-mAP@0.5). }
    \label{fig:da_res}
    \vspace{-0.3cm}
\end{figure}
in \uda, we conducted two more sets of experiments, in addition to the cross-dataset baseline described above and Table~\ref{tab:combine}. 
The experiment was set up in the following manner: 
%the source domain was 
\road was chosen as source domain and \roadwaymo as target domain; evaluation was performed on target domain's validation set.
Whenever a model is trained on the source domain with labels it is referred to as ``Lower bound'';
when a model is trained on the target domain with labels it is referred to as ``Upper bound''.
\uda is the training setup in 
which a model is trained on the source domain, with its labels, \emph{and} on the target domain's training set, but without labels.

Figure \ref{fig:da_res} illustrates how the results 
obtained using the Reverse Gradient (RevGrad) 
method (see Section~\ref{sec:dom_adapt_baseline}) fall within the range delimited by the Lower and Upper bounds as defined above. 
The closeness of the baseline results to the Lower Bound indicates the potential of the \roadpp family as a challenging real-world benchmark for developing and evaluating domain adaptation algorithms. 
Note that this is a real dataset-to-real dataset \uda setup, which makes it even more challenging than the more commonly considered %usually studied 
synthetic-to-real setups in semantic segmentation~\citep{hoyer2022daformer}.  
% One notable factor that contributes to the challenge posed by \roadwaymo is its substantial four-fold increase in the number of agents per frame compared to ROAD.
 
\begin{table}[t]
  \centering
  \setlength{\tabcolsep}{4pt}
  \caption{Baseline \textit{vs.} t-norm-based loss models, using I3D. Frame-level (mAP $\%$) are calculated on \roadwaymo val/test sets.}
  % \vspace{-3mm}
  {\footnotesize
  \scalebox{0.99}{
   \begin{tabular}{lcccc}
    \toprule
    Model &   Agents & Actions & Locations   \\     \midrule
    I3D   & 19.05 / 19.61 & 14.80 / 16.00 & 15.31 / 15.72  \\
    + G\"odel & \textbf{19.70} / \textbf{19.96} & \textbf{15.59} / 15.66 & \textbf{16.03} / \textbf{16.89}    \\
    + \L{}ukasiewicz & 18.07 / 19.89 & 15.13 / \textbf{16.13} & 15.37 / 15.84   \\
    % \midrule
    
    % Light-SlowFast  
    %                 & 16.01/\textbf{17.90} & 12.62/13.97 & 12.56/13.07   \\
    % + G\"odel   
    %                 & \textbf{16.83}/17.85 & \textbf{13.40}/\textbf{14.59}  & \textbf{14.10}/\textbf{14.07}     \\
    % + \L{}ukasiewicz & 15.35/17.15 & 12.62/13.94 & 12.33/12.69  \\
   \bottomrule
    \end{tabular}
  }
  }
\vspace{-2mm}
    \label{tab:t-norms_Roadwaymoframe-avg-results}
\end{table}

\subsection{Neuro-symbolic Results}

% Finally, in order to test whether logical requirements capturing background knowledge can help the neural networks learn better during training, we incorporated the requirements into a new t-norm-based loss term. 
Finally, we experimented with two types of t-norms (G\"{o}del and \L{}ukasiewicz) to compute to what extent the predictions satisfy the provided requirements, similarly to \roadr~\citep{giunchiglia2022road}.
% Additionally, we only applied the constraints on the bounding boxes for which the neural network assigned a confidence probability of over 0.5 that they contain an object.
% In this way, we avoided applying constraints such as \textit{there must be agent} to every bounding box, which would force the network to increase the score of one of the agent labels for bounding boxes unlikely to contain an agent.
% For our neuro-symbolic experiments, we used a lightweight backbone and focused only on the agents, actions and locations detection tasks (thus excluding the duplex and event detection tasks), where we could directly apply our constraints.\\
% We focused only on the agents, actions and locations detection tasks and excluded the duplex and event detection tasks---which would require a different set of constraints than the one the constraints   where we could directly apply our constraints.\\
%%%%%%%%%%%%%%%%%%%%%%%%%%%%%%%%%%%%%%%%%%%%%%%%%%%%%%%%%%%
As our requirements have been written specifically for the agents, actions and locations detection, we eliminated the duplex and triplet detection tasks (originally considered in ROAD) from this set of experiments.

The highest gains achieved with our method are reported for the location detection task when using the G\"{o}del t-norm, as shown in Table \ref{tab:t-norms_Roadwaymoframe-avg-results}, for the I3D model. Similar gains can be observed for SlowFast models, and can be consulted in supplementary material.
% and shows that using neuro-symbolic methods can lead to improved model performance and opens up further research directions for using domain knowledge during training in complex perception tasks, such as action and location detection, and real-world scenarios.
Whilst a modest improvement, this comes at no additional cost in terms of labelling or data acquisition, provided that the annotations have been logically cross-checked.
 However, regarding computational overhead, we note that enforcing such a large number of logical constraints makes the training process approximately $2.75$ times slower compared to the unconstrained baseline.
% Our method provides an essentially free performance enhancement and opens up further research directions for using domain knowledge during training in complex perception tasks %, such as action and location detection, 
% and real-world scenarios.
Despite this increase in training duration, our method provides a performance enhancement without requiring extra human  effort for image annotation, while also reducing the risk of making predictions that violate safety requirements.
This  opens up further research directions for using domain knowledge during training in complex perception tasks and real-world scenarios.

% an  performance enhancement and opens up further research directions for using domain knowledge during training in complex perception tasks %, such as action and location detection, 
% and real-world scenarios.

% \AB{@ Mihaela: Can we add something like "Whilst a modest improvement, this comes at no additional cost in terms of labelling or data aquisition, and thus provides an essentially free performance enhancement?" OR "Whilst a modest improvement, a safety-critical example of an incorrect detection that was prevented was XXX";} 

\section{Conclusions} % ANDY

\label{sec:conclusion}
%Autonomous vehicle perception must advance beyond object, into the ability to recognise and predict the future actions of road users - wherever, and whatever weather in which the vehicle may operate. 
%The few existing datasets for this purpose are small, and/or do not provide the comprehensive labelling required to enable the development and testing of models aimed at tackling this challenge.

The \roadwaymo dataset presented in this paper has been designed to enable 
the development of 
autonomous vehicle perception systems capable of advancing beyond object detection, and explore the ability to truly \textit{understand} what is happening around the vehicle. 
\roadwaymo is a large-scale dataset for activity and event detection and prediction---encompassing approximately $200k$ comprehensively labelled driving scenarios, spanning 4 different US cities, in 6 different (labelled) weather conditions. The integrity of the dataset has been confirmed and enhanced using a novel context-aware automated validation approach, allowing the identification and correction of thousands of potential human labelling errors.
This opens the door to tackling challenging perception problems including ``present time" tasks such as event, agent and action detection, as well as ``future time" prediction tasks such as action anticipation, trajectory prediction and so on.
Detailed baselines have been provided for the first set of tasks. 

As the (US) \roadwaymo is fully compatible with the original (UK) ROAD dataset, pairing the two datasets together forms the basis for a new ROAD++ family of datasets usable for the development and real-world testing of domain adaptation techniques - for which a baseline is also provided.
Finally, as the dataset comes with commonsense logical requirements, it can be used to develop and test neuro-symbolic methods for enhancing the neural networks' predictions. %- by allowing them access to domain knowledge during training. 
A baseline is also provided for this purpose.

\roadwaymo has been shown to present a significant challenge compared to existing datasets for this purpose, highlighting the scale of the work on AV perception still required to enable safe real-world driving. The ROAD++ family of datasets and challenges provides the perfect playground to enable the wider community to develop and push the boundaries of new approaches and methods for action and event detection in autonomous driving, and domain adaptation between countries.

% {\color{red}
\paragraph{Dataset Maintenance.}
ROAD-Waymo is released with semantic versioning. Annotation issues and bug reports are handled through a public GitHub~\footnote{\url{https://github.com/salmank255/Road-waymo-dataset}}. Future releases will include additional modalities as well as expanded geographic coverage.
% }

%\textbf{Limitations and future work}.
\textbf{Future work}.The dataset in its present form comes with data streams in other modalities (\eg LIDAR and GPS attached to each video), providing the potential for future expansion of the labelling into true multi-modal labels.
Another intriguing possibility is to project the road event annotation into the 3D space in the form of 3D detections,
%One could project our annotation in 3D for most of the videos 
where LIDAR and camera calibration is available.
% {\color{red}This IS in place for Waymo, right?}
% Every effort has been made to make video compatible, but the rest of the modalities isn't
In terms of results, the performance of the baselines provided is low for long-tail classes. We only used focal-loss~\citep{lin2017focal} to handle this problem without any explicit class balancing efforts, so more work needs to be done there. 

% Finally, we have plans to extend the \roadpp framework to additional domains and countries, such as the UAE.

% ... so there is a lot of work to be done...
% Long tail problem 

\section*{Acknowledgements}

This project has received funding from the European Union’s Horizon 2020 research and innovation programme, under grant agreement No. 964505 (E-pi). The project has also received funding from the Leverhulme Trust, Research Project Grant RPG-2019-243.
\section*{Broader Impact}

The work presented in this paper directly relates to ensuring safety in autonomous driving through an improved understanding of humans, but also of human-made environments. 
Considering not only pedestrian behaviour actions but also actions performed by humans as drivers of various types of vehicles, ROAD++ is part of an effort of shifting the paradigm from actions performed by human bodies to events caused by agents \cite{singh2022road}. %, which is crucial in real-world autonomous driving.

% The work presented here is intended at increasing the human-level situation awareness as a step towards attaining safe and reliable urban autonomous driving and reducing the likelihood of \href{https://www.tesladeaths.com/}{deadly accidents}, many of which are often caused by faults at the level of mere object recognition, thus highlighting the challenge of tackling more complex tasks such as detecting events or predicting intention.

% Autonomous driving systems inherently run the risk of causing accidents, some of which being \href{https://www.tesladeaths.com/}{deadly} and often caused by faults at the level of mere object recognition. 
Autonomous driving systems are inherently paired with the prospect of causing accidents, some of which being \href{https://www.tesladeaths.com/}{deadly} and often the effect of faults at the level of mere object recognition.
To this end, the work presented here highlights the challenge of tackling more complex tasks (such as detecting events) and is intended at drawing the researchers' attention towards the need of incorporating human-level situation awareness into self-driving systems.
One advantage we envision from this shift of paradigm is its potential to increase the public's confidence in autonomous vehicles by demonstrating different active research directions for real-world autonomous driving.

As a negative impact of the discussion presented in this paper, we discern that collecting data towards achieving an improved understanding of humans and their actions could lead to malicious parties attempting to profile the behaviour of individuals, thus highlighting possible risks of inadequate protection of privacy and the need of mitigating such dangers, e.g. by regulating the data collection.

% \AB{Need to discuss this section}
% Authors are required to include a statement of the broader impact of their work, including its ethical aspects and future societal consequences. 
% Authors should discuss both positive and negative outcomes, if any. For instance, authors should discuss a) 
% who may benefit from this research, b) who may be put at disadvantage from this research, c) what are the consequences of failure of the system, and d) whether the task/method leverages
% biases in the data. If authors believe this is not applicable to them, authors can simply state this.

% Use unnumbered first level headings for this section, which should go at the end of the paper. {\bf Note that this section does not count towards the eight pages of content that are allowed.}

% \begin{ack}
% Use unnumbered first level headings for the acknowledgments. All acknowledgments
% More information about this disclosure can be found at: \url{https://neurips.cc/Conferences/2023/PaperInformation/FundingDisclosure}.

% Do {\bf not} include this section in the anonymized submission, only in the final paper. You can use the \texttt{ack} environment provided in the style file to autmoatically hide this section in the anonymized submission.
% \end{ack}

\balance
\bibliographystyle{IEEEtran}
\bibliography{IEEEtran}

\end{document}